\documentclass{applemlr}

\usepackage{hyperref}
\usepackage{url}
\usepackage{amsthm}
\usepackage{hyperref}

\usepackage{fontawesome5}
\usepackage[smartEllipses]{markdown}

\usepackage[most]{tcolorbox}
\usepackage{wrapfig}
\usepackage{multirow}
\usepackage{graphicx,float}
\usepackage{threeparttable}
\usepackage[ruled,vlined]{algorithm2e}
\usepackage{colortbl}
\usepackage{color}
\usepackage{multirow}
\usepackage{tabularx}
\usepackage{float}
\usepackage{graphicx}
\usepackage{booktabs}
\usepackage{arydshln}
\usepackage{enumitem}
\usepackage{wrapfig}
\usepackage{caption}
\usepackage{graphicx}
\usepackage{makecell}
\usepackage{float} 
\usepackage{mathtools}
\usepackage{bbding}
\usepackage{makecell}
\usepackage{tabularx}
\usepackage{amssymb,mathrsfs,amsmath}
\usepackage{pifont}
\usepackage{bbm}
\usepackage{hhline}
\usepackage{boldline}
\usepackage{lmodern}

\usepackage{listings}

\newcommand{\eg}{\textit{e.g.}}
\newcommand{\ie}{\textit{i.e.}}

\newcommand{\ourmethod}{{\fontfamily{lmtt}\selectfont \textbf{WorldDraft-Primer}}\xspace}
\newcommand{\ourbench}{{\fontfamily{lmtt}\selectfont \textbf{WorldDraft}}\xspace}

\newcommand{\hlfirst}[1]{\colorbox{cyan!24}{#1}}  
\newcommand{\hlsecond}[1]{\colorbox{cyan!10}{#1}}

\definecolor{newgray}{gray}{0.4}

\usepackage[table,xcdraw,usenames,dvipsnames]{xcolor}
\usepackage{cleveref}
\SetKwInOut{Input}{Input}\SetKwInOut{Output}{Output}
\usepackage{twemojis}

\definecolor{headingpurple}{HTML}{7030A0}

\hypersetup{
    colorlinks=true,
    linkcolor=magenta,
    citecolor=cyan,
    filecolor=magenta,      
    urlcolor=magenta,
    }

\usepackage[most]{tcolorbox}

\definecolor{WDAccent}{HTML}{465F79}
\definecolor{WDFill}{HTML}{F7F9FC}
\definecolor{WDBorder}{HTML}{B9C7D6}

\tcbuselibrary{skins,breakable}

\newtcolorbox{coveragebox}[1][]{
    enhanced,
    breakable, % 新增：允许自动跨页
    colback=WDFill,
    colframe=WDBorder,
    boxrule=0.55pt,
    arc=1.8mm,
    outer arc=1.8mm,
    boxsep=0pt,
    left=8pt,
    right=8pt,
    top=12pt,
    bottom=10pt,
    before skip=10pt,
    after skip=8pt,
    title={World Coverage},
    fonttitle=\bfseries,
    coltitle=WDAccent,
    attach boxed title to top left={
        xshift=7pt,
        yshift=-\tcboxedtitleheight/2
    },
    boxed title style={
        colback=white,
        colframe=white,
        boxrule=0pt,
        arc=0pt,
        boxsep=0pt,
        left=4pt,
        right=4pt,
        top=1pt,
        bottom=1pt
    }, % 这里加逗号
    #1 % 放在最后，允许覆盖前面的默认设置
}

\newcommand{\omniweavinglogo}{\raisebox{-0.15em}{\includegraphics[height=1em]{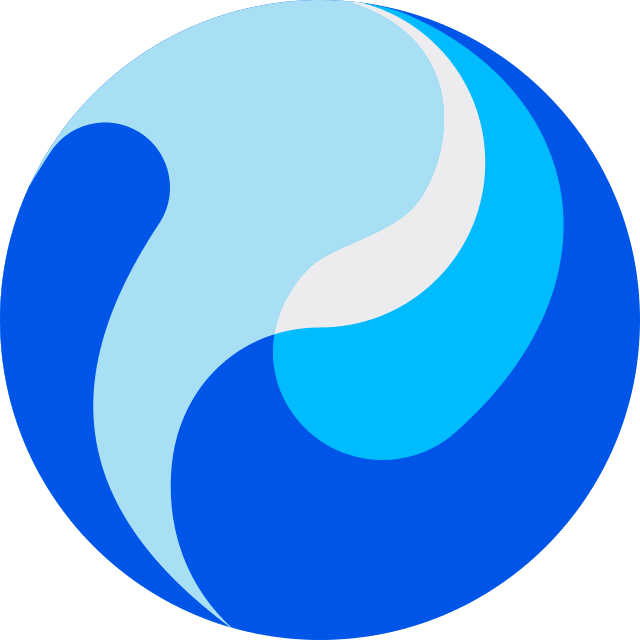}}}
\newcommand{\univideologo}{\raisebox{-0.15em}{\includegraphics[height=1em]{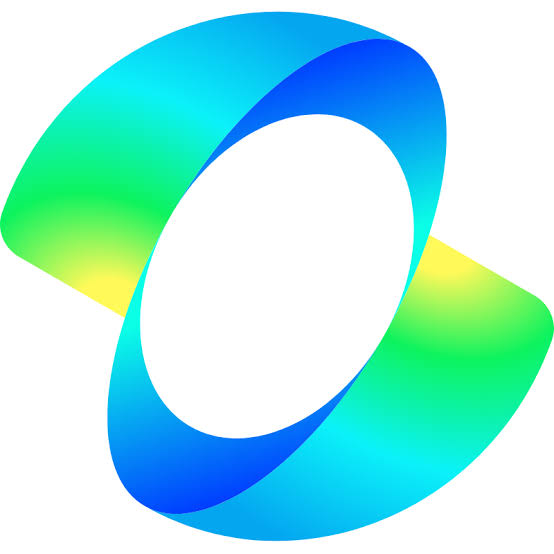}}}
\newcommand{\vinologo}{\raisebox{-0.15em}{\includegraphics[height=1em]{logo/kling.jpeg}}}
\newcommand{\wanlogo}{\raisebox{-0.15em}{\includegraphics[height=1em]{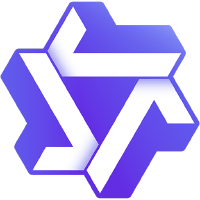}}}
\newcommand{\ctorlogo}{\raisebox{-0.15em}{\includegraphics[height=1em]{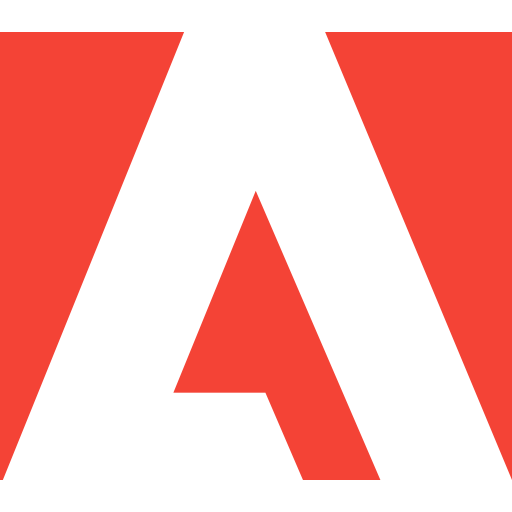}}}
\newcommand{\berninilogo}{\raisebox{-0.15em}{\includegraphics[height=1em]{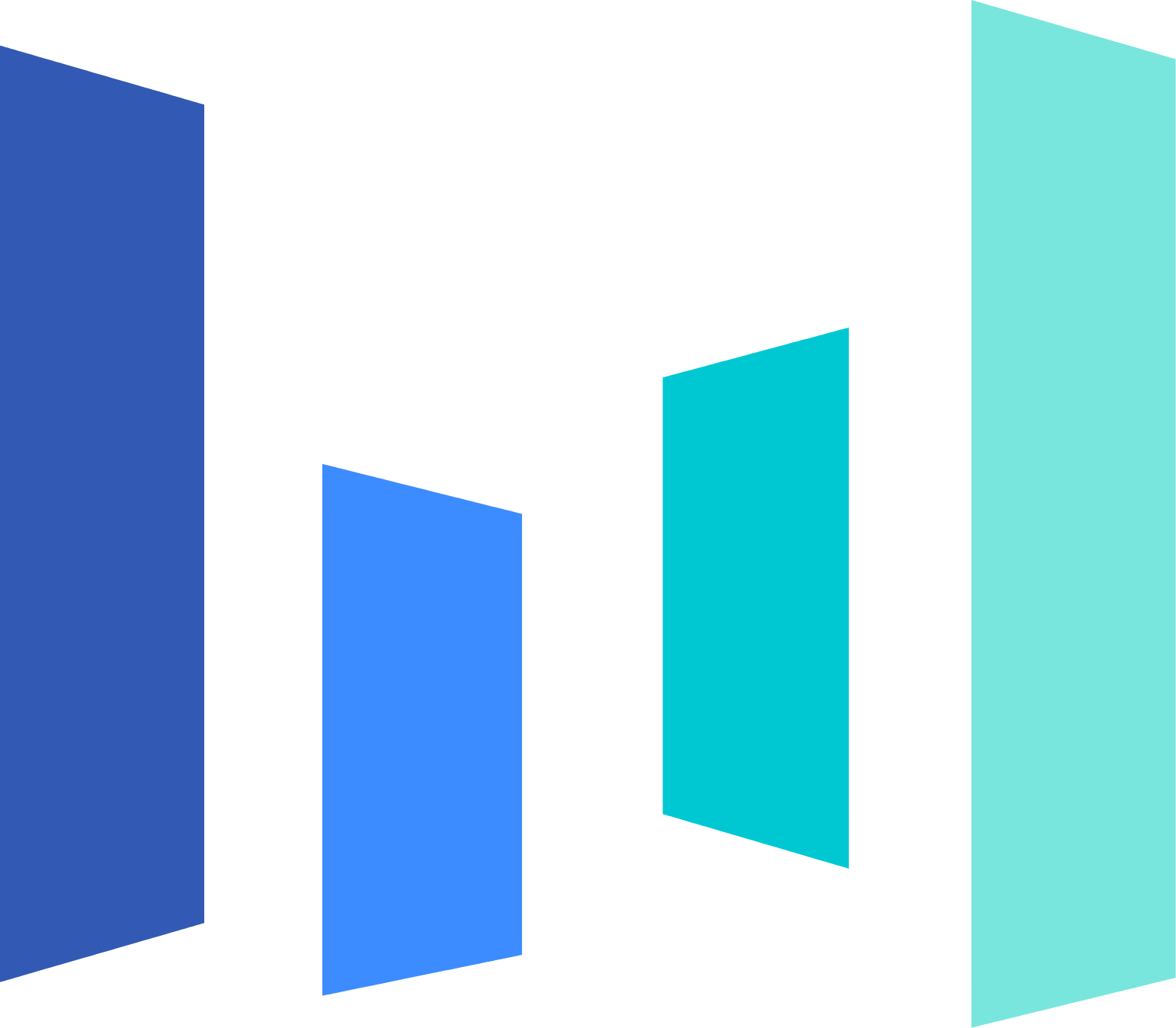}}}
\newcommand{\seedancelogo}{\raisebox{-0.15em}{\includegraphics[height=1em]{logo/bytedance-7aa9dfbc.png}}}
\newcommand{\lynnreallogo}{\raisebox{-0.15em}{\includegraphics[height=1em]{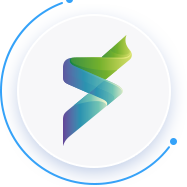}}}
\newcommand{\minimaxlogo}{\raisebox{-0.15em}{\includegraphics[height=1em]{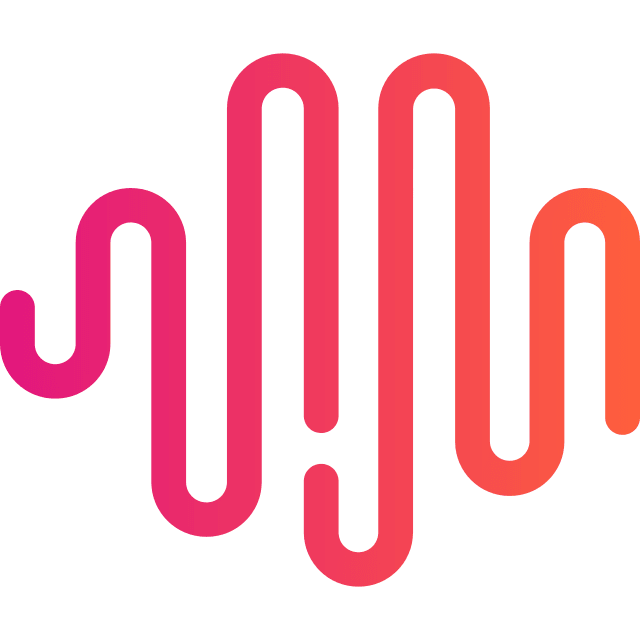}}}
\newcommand{\solarlogo}{\raisebox{-0.15em}{\includegraphics[height=1em]{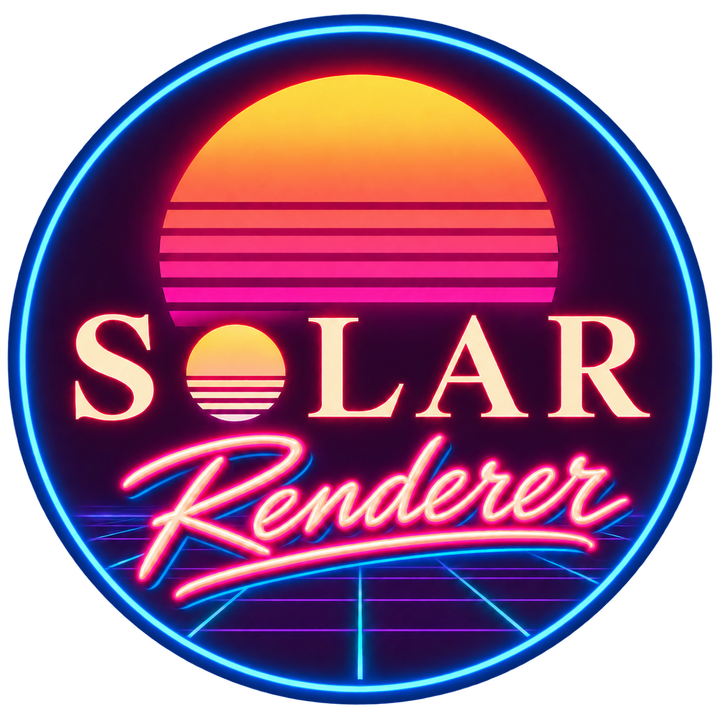}}}

\title{WorldDraft: If Code Builds the World, Can Video Models Just Render It?}

\author[1*]{Zhiyu Hou}
\author[2*]{Wen-Jie Shu}
\author[3*\dagger]{Harold H. Chen}
\author[4]{Weilin Ruan}
\author[5]{Chenfei Liao}
\author[3]{Liuzhou Zhang}
\author[6]{Disen Lan}
\author[7]{Xiaogang Xu}
\author[2]{Qi Chen}
\author[8]{Ser-Nam Lim}
\affiliation[1]{SUSTech}
\affiliation[2]{ZODA}
\affiliation[3]{HKUST}
\affiliation[4]{CUHK}
\affiliation[5]{HKUST(GZ)}
\affiliation[6]{FDU}
\affiliation[7]{ZJU}
\affiliation[8]{UCF}
\contribution[*]{Equal Contribution}
\contribution[\dagger]{Project Lead}
\abstract{Multimodal coding agents decouple 4D world construction from visual generation, repurposing video models from world generators toward \textit{generative renderers}.
This division of labor raises a fundamental question: \textbf{if code builds the world, can video models just render it?}
A reliable generative renderer hinges on two complementary criteria: (\textbf{\textit{i}})~\textit{world preservation}, retaining prescribed 4D kinematics and camera paths; and (\textbf{\textit{ii}})~\textit{visual realization}, persistently binding text-specified target attributes over time.
To evaluate this capability, we introduce \ourbench, a benchmark of $432$ strictly paired test cases across $48$ executable 4D world families spanning six dynamic challenge categories.
Under a $3 \times 3$ factorial design, \ourbench orthogonally crosses \textbf{draft granularity} (abstract proxy to previs) and \textbf{appearance alignment} (neutral to conflicting) under identical dynamics.
Extensive evaluations across $12$ representative models reveal that:
(I)~world preservation and visual realization remain in stark operational tension;
(II)~richer draft scaffolding anchors structure but paradoxically impairs appearance fidelity; and
(III)~conflicting source cues trigger severe source leakage (up to $64\%$), exposing fragile cross-modal disentanglement.
To mitigate these bottlenecks, we propose \ourmethod, a plug-and-play visual prompting strategy that consistently improves both objectives across frozen video renderers.
Together, \ourbench and \ourmethod lay a foundation for bridging programmatic 4D simulation with video generation, advancing physically grounded generative rendering.}

\metadata[{\faGlobe\ Project}]{\url{https://haroldchen19.github.io/WorldDraft-site/}}
\metadata[{\faGithub\ Github}]{\url{https://github.com/KevinHuge/WorldDraft}}

\begin{document}
\maketitle

\begin{figure*}[!h]
\centering
% \vspace{-0.4em}
\includegraphics[width=\linewidth]{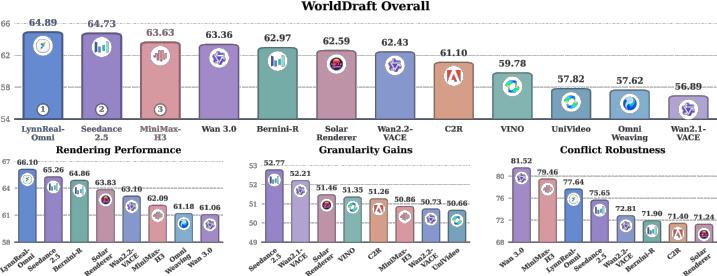}
\vspace{-1.9em}
\caption{Performance overview of $12$ video models on \ourbench across $3$ dimensions.}
\label{fig:overall}
\vspace{-0.4em}
\end{figure*}

% \begin{figure*}[!t]
% \centering
% % \vspace{-0.4em}
% \includegraphics[width=0.98\linewidth]{figures/Figure_overall.pdf}
% % \vspace{-2em}
% \vspace{-0.8em}
% \caption{Performance overview of $12$ video models on \ourbench across $3$ dimensions.}
% \label{fig:overall}
% \vspace{-0.8em}
% \end{figure*}

% \vspace{-1em}
\section{Introduction}
\label{sec:intro}

Video generation has achieved remarkable progress in visual realism and temporal coherence, yet generating complex, controllable dynamic scenes remains challenging \citep{wan2025wan, agarwal2025cosmos, li2026video, hacohen2024ltx}. Since end-to-end models must implicitly infer 4D world evolution from brief text while simultaneously generating pixels, high visual quality often comes at the expense of physical and spatiotemporal reliability. Recently, advances in \textit{multimodal coding agents} have enabled an alternative paradigm that decouples world construction from visual generation \citep{li2026videococo, wang2026code, liang2026codevideomodel, zhang2026gs}. By programmatically controlling engines such as Blender \citep{BlenderOnlineCommunity2026} or MuJoCo \citep{todorov2012mujoco}, coding agents can explicitly specify scene layouts, object trajectories, and physical interactions as executable simulation drafts, leaving video generation models to transform these low-fidelity drafts into photorealistic videos.

This emerging division of labor explores repurposing video models from \textit{\textbf{world generators}} toward \textit{\textbf{generative renderers}}, raising a fundamental question: \textit{if code builds the world, can video models just render it?} We thus argue that a reliable generative renderer hinges on fulfilling two complementary objectives: \ding{182}~\textbf{world preservation}, which demands faithfully retaining the scene geometry, object dynamics, event timing, and camera motion prescribed by the draft without rewriting the underlying world; and \ding{183}~\textbf{visual realization}, which requires accurately generating text-specified target attributes (\eg, color, material) while persistently binding them to the corresponding 4D entities across motion, occlusion, and viewpoint changes.

Formally, generative rendering operates as a conditional video-to-video (V2V) transformation guided by a structural draft and a target text prompt, yet existing benchmarks only partially address its requirements. Early V2V benchmarks \citep{li2025five, liu2026v2v} focus on editing real-world videos that already possess rich source textures, whereas recent programmable-world evaluation \citep{huang2026rowbench} primarily examines whether generated videos follow program-logged states and interaction events. Crucially, achieving \textit{world preservation} does not guarantee faithful \textit{visual realization}: models frequently retain coarse draft dynamics while failing to generate or temporally bind target appearance, or conversely fulfill appearance prompts at the cost of altering the underlying 4D world. This gap motivates the need for a new benchmark that jointly evaluates both objectives to systematically characterize the capability boundaries of generative renderers.

\begin{figure*}[!t]
\centering
% \vspace{-0.4em}
\includegraphics[width=\linewidth]{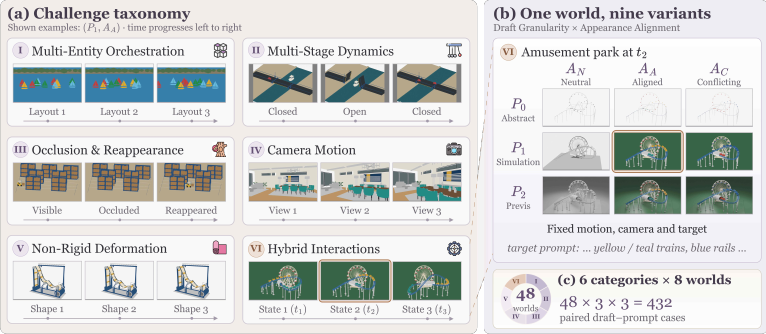}
\vspace{-1.9em}
\caption{Overview of \ourbench. (\textbf{\textit{a}})~Six challenge categories across $48$ executable 4D worlds (shown at $P_1, A_A$). (\textbf{\textit{b}})~$3 \times 3$ factorial grid crossing draft granularity ($P_0 \to P_2$) and appearance alignment ($A_N, A_A, A_C$). (\textbf{\textit{c}})~Full benchmark totaling $48 \times 3 \times 3 = 432$ cases.}
\label{fig:teaser}
\vspace{-0.4em}
\end{figure*}

In this work, we propose \ourbench, a systematic benchmark designed to evaluate video models as draft-to-video generative renderers through three progressive questions (\Cref{fig:teaser}). First, \ding{192}~\textit{can video models render target appearances without corrupting complex 4D worlds?} To test this foundational capability, we curate $48$ executable 4D world families spanning \ding{168}~\textbf{six challenge categories}: multi-entity orchestration, multi-stage dynamics, occlusion and reappearance, camera motion, non-rigid deformation, and hybrid interactions. Grounded in explicit simulator references, \ourbench jointly measures how faithfully a model preserves the prescribed 4D structure and how accurately it realizes and persistently binds target visual attributes over time. Building upon these base worlds, we further probe the capability boundaries of generative renderers along two complementary axes: \ding{193}~\textit{how much information must a draft provide to support reliable rendering?} We then instantiate each world across \ding{169}~\textbf{three draft granularity levels} (ranging from abstract proxies to detailed previews) to quantify minimal input requirements and the gains from richer structural and visual cues; and \ding{194}~\textit{can models disentangle the world structure to preserve from the source appearance to overwrite?} We further pair each draft with \ding{171}~\textbf{three appearance alignment levels} (neutral, aligned, and conflicting) under a shared target prompt to evaluate selective adherence to structural versus textual conditions. Crossing both axes across all $48$ world families yields a unified $3\times 3$ factorial design of $432$ strictly paired test cases.

Through extensive evaluation on \ourbench, we observe that existing video models struggle to jointly preserve underlying 4D worlds and realize persistent target appearances, especially under coarse draft granularity or conflicting source-target cues (see \Cref{fig:overall}). Motivated by these findings, we further propose \ourmethod, a plug-and-play visual prompting strategy designed to enhance frozen video renderers. Unlike existing adaptation pipelines that rely on costly retraining or privileged simulator access \citep{li2026videococo, gomez2026coarse}, \ourmethod operates solely on the input draft and target prompt. Specifically, it injects subtle visual cues into the first draft frame via two complementary strategies: (\textbf{\textit{i}})~\textbf{object emphasis}, which highlights key or easily overlooked entities to reinforce \textit{world preservation}; and (\textbf{\textit{ii}})~\textbf{attribute cueing}, which encodes target colors, textures, and shading cues around entities and supporting surfaces to guide \textit{visual realization}. Experiments show that \ourmethod consistently boosts performance across diverse models, serving as both an analytical companion to \ourbench and a practical solution for improving draft-to-video generative rendering.

To summarize, this work contributes threefold: 
\begin{itemize}[leftmargin=*]
    \item We propose \ourbench, a systematic benchmark for draft-to-video generative rendering comprising $432$ strictly paired test cases across $48$ executable 4D world families, organized across six challenge categories and a $3\times 3$ factorial grid of draft granularity and appearance alignment levels.
    \item Through extensive experiments across $12$ representative video models, our findings highlight that: (\textbf{\textit{i}})~world preservation and visual realization remain in stark operational tension; (\textbf{\textit{ii}})~finer draft granularity anchors structure but paradoxically impairs appearance fidelity; and (\textbf{\textit{iii}})~conflicting source cues cause severe source leakage, exposing fragile source-target disentanglement.
    \item We further introduce \ourmethod, a simple yet effective plug-and-play visual prompting strategy that injects object emphasis and attribute cues into the first draft frame, consistently boosting both world preservation and visual realization across frozen video renderers.
\end{itemize}

\section{Related Work}
% \vspace{-0.4em}

\paragraph{Code-for-Vision.}
Emerging works explore executable code as an intermediate representation for 3D reconstruction and scene synthesis \citep{yin2026vision, wang20263d, huang2024blenderalchemy}, driving graphics engines to explicitly control geometry, camera parameters, and lighting. Extending this paradigm to dynamic environments, frameworks such as GS-Agent \citep{zhang2026gs} and Code as Worlds \citep{wang2026code} describe 4D physical evolution in code, relying on simulators to govern deterministic trajectories and interaction rules. This approach establishes a modular division of labor: systems like VideoCoCo \citep{li2026videococo} and Code Video Model \citep{liang2026codevideomodel} also employ coding agents to construct simulation drafts, leaving video generation models to synthesize photorealistic outputs. While prior research primarily targets upstream world construction, \ourbench focuses on the downstream rendering stage, evaluating whether video models can faithfully realize target appearances without corrupting the underlying executable world.

\vspace{-0.8em}
\paragraph{Evaluation of Video-to-Video.}
V2V evaluation has traditionally focused on text-guided editing, measuring instruction adherence, temporal consistency, and background preservation on natural videos \citep{chen2025editboard,li2025five,liu2026v2v,gao2026vefx}. With the rise of programmatic generation, recent benchmarks such as PROWBench \citep{huang2026rowbench} check behavioral fidelity and event timelines against engine-logged states while evaluating visual quality through holistic metrics such as CLIP. However, such holistic evaluations overlook whether models can faithfully realize fine-grained target attributes and persistently bind them across dynamic 4D processes. In this work, \ourbench goes a step further by investigating how varying input conditions govern the interplay between world preservation and appearance realization.

\section{WorldDraft}

\subsection{Task Formulation}
\label{sec:task_formulation}

\paragraph{Generative Rendering.} 
Programmatic video generation decouples physical reasoning from visual generation: upstream engines explicitly govern scene kinematics, leaving downstream video models to act as renderers. Formally, an upstream simulator executes an underlying 4D world $\mathcal{W}$ to yield a draft video $\mathcal{D} = \{\mathbf{I}_t^{\text{src}}\}_{t=1}^T\in \mathbb{R}^{T \times H \times W \times 3}$. Given $\mathcal{D}$ and a target prompt $\mathbf{p}$, a generative renderer $\mathcal{G}$ synthesizes the target video:
\setlength\abovedisplayskip{3.4pt}
\setlength\belowdisplayskip{3.4pt}
\begin{equation}
    \hat{\mathcal{V}} = \mathcal{G}(\mathcal{D}, \mathbf{p}),
    \label{eq:rendering_formulation}
\end{equation}
where $\mathcal{D}$ provides the 4D spatiotemporal scaffold (\eg, geometry, kinematics, camera), and $\mathbf{p}$ dictates target visual attributes (\eg, color, material, lighting). At inference, $\mathcal{G}$ strictly accesses $(\mathcal{D}, \mathbf{p})$, with $\mathcal{W}$ reserved solely for draft generation and ground-truth evaluation references.

\vspace{-0.8em}
\paragraph{Dual Criteria.} 
To determine whether $\mathcal{G}$ faithfully executes this conditional transformation, the correctness of generative rendering hinges on two criteria:
\begin{enumerate}[leftmargin=1.3em, itemsep=1pt, topsep=1pt]
    \item[\ding{111}] \textbf{World Preservation}, requiring $\hat{\mathcal{V}}$ to faithfully preserve the prescribed 4D kinematics, entity identities, event timelines, and camera motion in $\mathcal{D}$ without prompt-countermanded structural hallucinations.
    \item[\ding{111}] \textbf{Visual Realization}, requiring $\hat{\mathcal{V}}$ to accurately instantiate target attributes from $\mathbf{p}$ and persistently bind them to corresponding entities across dynamic motion, occlusions, and viewpoint changes.
\end{enumerate}
Generative rendering thus demands that $\mathcal{G}$ disentangle the \textit{world structure to preserve} from the \textit{source appearance to overwrite}: a capability fundamentally governed by the draft's input condition.

\subsection{Executable World Families}
\label{sec:world_families}

\paragraph{Challenge Taxonomy.}
To evaluate renderers beyond trivial rigid motions, candidate worlds should enforce deterministic physical constraints for \textit{world preservation} while imposing dynamic stressors for \textit{visual realization}. Guided by these principles, we first establish $48$ executable world families across \textbf{six primary categories} (\Cref{fig:teaser}):
% \vspace{-0.4em}
\vspace{0.4em}
\begin{coveragebox}
\noindent
{\color{WDAccent}\textbf{I.}\enspace
\textbf{Multi-Entity Orchestration:}}
Preserving multi-entity spatial arrangements and coordination, challenging the renderer to bind distinct target attributes without identity cross-contamination.
\par\addvspace{4pt}

\noindent
{\color{WDAccent}\textbf{II.}\enspace
\textbf{Multi-Stage Dynamics:}}
Retaining causal event ordering and phase transitions, challenging attribute persistence across state-changing boundaries.
\par\addvspace{4pt}

\noindent
{\color{WDAccent}\textbf{III.}\enspace
\textbf{Occlusion \& Reappearance:}}
Maintaining physical occlusion orders and reappearance trajectories, challenging the renderer to re-identify and restore the exact target appearance post-emergence.
\par\addvspace{4pt}

\noindent
{\color{WDAccent}\textbf{IV.}\enspace
\textbf{Camera Motion:}}
Retaining 6-DoF camera trajectories and parallax geometry, demanding strict multi-view appearance consistency under perspective shifts.
\par\addvspace{4pt}

\noindent
{\color{WDAccent}\textbf{V.}\enspace
\textbf{Non-Rigid Deformation:}}
Retaining continuous mesh topology and soft-body kinematics, requiring surface textures and materials to deform realistically without visual slipping or artifacts.
\par\addvspace{4pt}

\noindent
{\color{WDAccent}\textbf{VI.}\enspace
\textbf{Hybrid Interactions:}}
Retaining dynamic physical coupling across disparate mechanisms (\eg, collisions and force transfer), challenging identity and material retention under violent multi-body contact.
\end{coveragebox}
\vspace{-0.2em}
\noindent These categories serve as primary stressors rather than disjoint splits; each family is classified by its dominant challenge while incorporating secondary dynamics. In particular, \textit{multi-entity orchestration} focuses on cooperative spatial topology, whereas \textit{hybrid interactions} stresses dynamic physical coupling between contact processes.

\begin{figure*}[!t]
\centering
% \vspace{-0.4em}
\includegraphics[width=\linewidth]{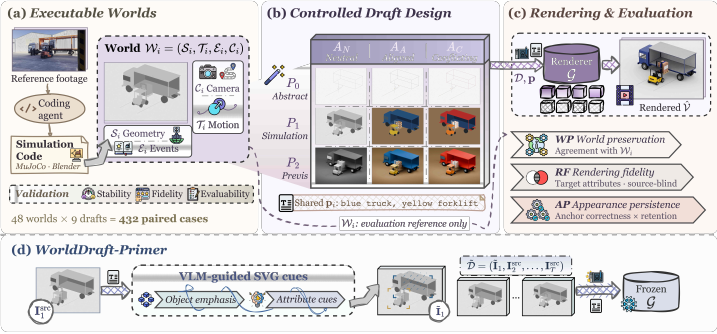}
\vspace{-2em}
% \vspace{-0.8em}
\caption{Overview of our proposed (\textbf{\textit{Top}}) \ourbench benchmark and (\textbf{\textit{Bottom}}) \ourmethod strategy.}
\label{fig:pipeline}
\vspace{-0.8em}
\end{figure*}

\vspace{-0.8em}
\paragraph{Construction and Validation.}
For each challenge, we first collect diverse real-world videos as dynamic blueprints, and employ multimodal coding agents (\ie, GPT-6-Astra~\citep{singh2025openai}) to translate their underlying physical interactions and camera paths into executable MuJoCo~\citep{todorov2012mujoco} or Blender~\citep{BlenderOnlineCommunity2026} scripts.
Crucially, the footage serves as world reference rather than a target for pixel-level reconstruction.
Execution produces deterministic 4D rollouts while programmatically logging the ground-truth state tuple:
\begin{equation}
\mathcal{W}_i = (\mathcal{S}_i, \mathcal{T}_i, \mathcal{E}_i, \mathcal{C}_i), 
\end{equation}
with 3D geometry $\mathcal{S}_i$, dense trajectories $\mathcal{T}_i$, event timestamps $\mathcal{E}_i$, and camera poses $\mathcal{C}_i$.
Each world passes a three-tiered admission audit: \ding{192}~numerical stability without divergence, \ding{193}~fidelity to the reference interaction, and \ding{194}~evaluative viability with unambiguous 4D binding stressors, freezing $48$ standardized families, as shown in \Cref{fig:pipeline}.

\subsection{Controlled Draft Design}
\label{sec:controlled_draft_design}

To further isolate whether rendering failures stem from insufficient structural support or deceptive appearance cues, we then freeze the 4D kinematics and prompt $\mathbf{p}_i$ for each $\mathcal{W}_i$, and further orthogonally vary drafts across two axes: \textit{draft granularity} and \textit{appearance alignment}.

\vspace{-0.8em}
\paragraph{Prompt Suite.}
A unified prompt $\mathbf{p}_i$ per family specifies $2\sim4$ verifiable attributes (\eg, color, material, lighting). To guarantee unbiased evaluation, $\mathbf{p}_i$ enforces: \ding{192}~\textit{observable verification} via unambiguous visual criteria; \ding{193}~\textit{alignment-invariant references} that index entities via spatial/semantic roles (\eg, ``leading cube'') rather than volatile appearance tags; and \ding{194}~\textit{structural neutrality}, strictly omitting trajectories or camera paths so text cannot compensate for draft sparsity. More examples are in \Cref{sec:appendix_case_demo}.

\vspace{-0.8em}
\paragraph{Draft Granularity ($P$).}
To investigate minimal input requirements, we instantiate each world across three monotonic information levels $P \in \{P_0, P_1, P_2\}$:
\begin{enumerate}[leftmargin=1.5em, itemsep=0pt, topsep=1pt]
    \item[\ding{182}] \textbf{Abstract ($P_0$):} Primitive geometries and flat shading; retains basic semantics and kinematics to test low-scaffold completion.
    \item[\ding{183}] \textbf{Simulation ($P_1$):} Standard physical meshes and basic lighting; serves as the canonical draft-to-video rendering operating point.
    \item[\ding{184}] \textbf{Previs ($P_2$):} Dense meshes, full context, and realistic shadows; quantifies headroom while retaining a synthetic-to-real gap.
\end{enumerate}
This axis establishes an \textit{information ladder} of structural completeness rather than cosmetic image filtering.

\vspace{-0.8em}
\paragraph{Appearance Alignment ($A$).}
Rich visual cues can either guide rendering or trigger severe source leakage. Therefore, to evaluate adherence to text over deceptive source cues, each granularity $P$ is paired with three appearance alignment states $A \in \{A_N, A_A, A_C\}$:
\begin{enumerate}[leftmargin=1.5em, itemsep=0pt, topsep=1pt]
    \item[\ding{182}] \textbf{Neutral ($A_N$):} Uninformative placeholder shaders (\eg, gray clay), relying solely on text for appearance.
    \item[\ding{183}] \textbf{Aligned ($A_A$):} Source appearances match $\mathbf{p}_i$ within the precision of the respective granularity level.
    \item[\ding{184}] \textbf{Conflicting ($A_C$):} Contradictory source cues (\eg, blue metal \textit{vs.} target red plastic) to probe source leakage.
\end{enumerate}
Variations affect strictly surface appearance without altering geometry $\mathcal{S}_i$, kinematics $\mathcal{T}_i$, or camera $\mathcal{C}_i$. 

Crossing both axes yields a unified $3 \times 3$ grid of $9$ paired inputs $\mathcal{D}_i^{(j,k)}$ for each world, totaling $48 \times 3 \times 3 = 432$ strictly paired test cases. Fixing alignment isolates the marginal gain of draft information ($\Delta_{01}, \Delta_{12}$), while fixing granularity isolates conflict sensitivity and source leakage.

\subsection{Evaluation Suite}
\label{sec:evaluation_protocol}

Standard video metrics (\eg, FVD \citep{unterthiner2018towards}) fail to accurately evaluate both world preservation and visual realization directly. We thus establish an analytical suite across three disentangled dimensions: \ding{192}~\textit{world preservation} ($\mathrm{WP}$), \ding{193}~\textit{rendering fidelity} ($\mathrm{RF}$), and \ding{194}~\textit{appearance persistence} ($\mathrm{AP}$).
% each evaluated independently against simulator references without collapsing into a monolithic score.
We report these dimensions separately to preserve diagnostic interpretability, with an additional composite score for overall comparison.

\vspace{-0.8em}
\paragraph{World Preservation ($\mathrm{WP}$).}
Grounded in $\mathcal{W}_i = (\mathcal{S}_i, \mathcal{T}_i, \mathcal{E}_i, \mathcal{C}_i)$, $\mathrm{WP}$ independently benchmarks: \ding{192} \textit{Entity \& Layout ($\mathcal{S}_i$)}, using Count Accuracy and trajectory-matched BBox IoU; \ding{193} \textit{Kinematics \& Events ($\mathcal{T}_i, \mathcal{E}_i$)}, via Matched Track Coverage, Interaction Success Rate (ISR), and Event Timing Coverage; and \ding{194} \textit{Camera ($\mathcal{C}_i$)}, via relative rotation and scale-normalized translation errors against the canonical draft. Metrics evaluate intended topology, ensuring $P_0$ is not penalized for lacking fine details only introduced in $P_2$. Details in \Cref{sec:appendix_rendering}.

\vspace{-0.8em}
\paragraph{Rendering Fidelity ($\mathrm{RF}$).}
Adherence to $\mathbf{p}_i$ is evaluated via attribute-level binary QA (Yes/No) across target colors, materials, and lighting, incorporating balanced affirmative/negated probes. Critically, \textit{QA judges receive only the target text and generated video, remaining strictly isolated from source drafts to prevent bias}. Standard perceptual scores (\eg, Image Quality, Aesthetic) are reported separately as non-gated references $\mathrm{VQ}$.

\vspace{-0.8em}
\paragraph{Appearance Persistence ($\mathrm{AP}$).}
Since maintaining an incorrect appearance over time is not always consistency, we therefore establish a \textit{correctness-gated} persistence metric:
\begin{equation}
    \mathrm{AP} = \frac{100}{|\mathcal{Q}_{\mathrm{AP}}|} \sum_{q \in \mathcal{Q}_{\mathrm{AP}}} c_q \cdot s_q,
    \label{eq:ap_metric}
\end{equation}
where $c_q \in \{0, 1\}$ verifies target attribute realization at an anchor frame, and $s_q \in \{0, 1\}$ checks whether the identical entity retains that attribute across subsequent timestamps. Credit is awarded if and only if both pass ($c_q \cdot s_q = 1$).

\vspace{-0.8em}
\paragraph{Grid Diagnostics and Aggregation.}
On the full $3\times3$ grid, we designate $(P_1, A_N)$ as the canonical operating point and derive two diagnostic quantities: (\textbf{\textit{i}}) \textit{Conflict Penalty}, $\mathrm{CP} = \mathrm{RF}(A_A) - \mathrm{RF}(A_C)$, measuring performance drop under conflicting cues, and (\textbf{\textit{ii}}) \textit{Source Leakage}, $\mathrm{SL} = N_{\text{source\_leaked}} / N_{\text{conflict}}$, quantifying the erroneous persistence of conflicting source attributes.

\section{WorldDraft-Primer}

The dual correctness criteria of \ourbench expose an inherent operational dilemma in generative rendering: \textit{how to reliably anchor target attributes without corrupting prescribed kinematics.}
% Extensive benchmark evaluation reveals that frozen renderers struggle with spatial attribute binding and entity omission under coarse drafts.
While prompt engineering might seem intuitive, text prompts are inherently spatially agnostic, frequently inducing attribute cross-contamination across multiple entities. Conversely, modifying all draft frames risks destroying the deterministic kinematics and inducing temporal flickering. Leveraging the fact that video diffusion models naturally propagate initial appearances via spatiotemporal attention \citep{fang2025video, qi2026mme}, we propose \ourmethod, a plug-and-play \textit{visual prompting} strategy. \ourmethod intervenes strictly on the initial frame $\mathbf{I}_1^{\text{src}}$, producing a cued frame $\tilde{\mathbf{I}}_1$:
\begin{equation}
    \tilde{\mathcal{D}} = (\tilde{\mathbf{I}}_1, \mathbf{I}_2^{\text{src}}, \dots, \mathbf{I}_T^{\text{src}}), \qquad \hat{\mathcal{V}} = \mathcal{G}(\tilde{\mathcal{D}}, \mathbf{p}),
    \label{eq:primer_pipeline}
\end{equation}
leaving the remaining $T-1$ draft frames untouched to preserve the exact physical evolution.

\vspace{-0.8em}
\paragraph{Object Emphasis for World Preservation.}
To prevent renderers from dropping small or low-contrast entities, \textit{object emphasis} introduces sparse, non-intrusive bounding cues. Guided by the prompt $\mathbf{p}$, an open-source VLM (\ie, Qwen3.5 \citep{yang2025qwen3}) identifies $K$ critical entities, while an open-vocabulary grounding model predicts their 2D bounding boxes $\mathcal{B} = \{\mathbf{b}_k\}_{k=1}^K$. For each box $\mathbf{b}_k$, we generate a binary mask $\mathbf{M}_k^{\text{corner}} \in \{0, 1\}^{H \times W}$ outlining minimalist corner brackets strictly around the entity periphery. The structural emphasis signal is formulated as:
\begin{equation}
    \Delta \mathbf{I}_{\text{OE}} = \sum_{k=1}^K \mathbf{M}_k^{\text{corner}} \odot \mathbf{C}_{\text{mono}},
    \label{eq:oe_formula}
\end{equation}
where $\mathbf{C}_{\text{mono}}$ is a neutral monochrome tone. Since $\mathbf{M}_k^{\text{corner}}$ occupies only narrow peripheral margins without covering entity interiors or physical contact surfaces, it acts purely as a spatial anchor without altering underlying geometry.

\vspace{-0.8em}
\paragraph{Attribute Cueing for Visual Realization.}
To steer appearance binding, \textit{attribute cueing} injects non-textual hints across two levels: (\textbf{\textit{i}}) \textit{entity cues} ($\Delta \mathbf{I}_{\text{ent}}$), which modulate peripheral brackets with target colors, simplified texture patterns, or coarse specular shading while leaving source interiors visible; and (\textbf{\textit{ii}}) \textit{environment cues} ($\Delta \mathbf{I}_{\text{env}}$), which introduce soft contact shadows, coarse ground textures, or subtle lighting gradients onto supporting surfaces. The composite initial frame $\tilde{\mathbf{I}}_1$ is formulated as:
\begin{equation}
    \tilde{\mathbf{I}}_1 = \mathbf{I}_1^{\text{src}} \odot (1 - \mathbf{M}_{\text{all}}) + \big(\Delta \mathbf{I}_{\text{OE}} + \Delta \mathbf{I}_{\text{ent}} + \Delta \mathbf{I}_{\text{env}}\big) \odot \mathbf{M}_{\text{all}},
    \label{eq:primer_synthesis}
\end{equation}
where $\mathbf{M}_{\text{all}} = \sum_k \mathbf{M}_k^{\text{corner}} + \mathbf{M}_{\text{env}}$ is the aggregate cue mask. Cues remain strictly localized to keep source-target conflicts ($A_C$) uncompromised. To avoid generative noise and pixel blurring from diffusion inpainting, all prompts are synthesized programmatically via SVG, ensuring sub-pixel precision and strictly bounded transparency.

\section{Experiments}

\subsection{Experimental Settings}

\paragraph{Models.} We benchmark $12$ representative video models on \ourbench, spanning three primary paradigms across $10$ open-weight architectures:
\ding{192}~\textit{Generative Renderers}: SolarRenderer-33B \citep{fang2026solarrenderer}, C2R-14B \citep{gomez2026coarse};
\ding{193}~\textit{Omni Models}: MiniMax-H3-33B \citep{minimax2025minimaxh3}, LynnReal-Omni-32B \citep{mao2026lynnreal}, OmniWeaving-8.3B \citep{pan2026omniweaving}, VINO-13B~\citep{chen2026vino}, UniVideo-13B \citep{wei2026univideo}; and
\ding{194}~\textit{Controllable Generators}: Bernini-R-14B \citep{team2026bernini}, Wan2.2-VACE-Fun-A14B \citep{jiang2025vace}, Wan2.1-VACE-14B \citep{jiang2025vace}.
We additionally evaluate two leading commercial models: Wan~3.0 \citep{wan2025wan} and Seedance~2.5 \citep{bytedance2025seedance25}.
All models take $(\mathcal{D}, \mathbf{p})$ as input, ensuring consistent conditioning.

\begingroup
\setlength{\tabcolsep}{1.6pt}
\begin{table*}[t]
\vspace{-0.6em}
\renewcommand{\arraystretch}{1.6}
\centering
\caption{Main results on \ourbench. We highlight the \hlfirst{best} and \hlsecond{second best} scores ($\uparrow/\downarrow$: higher/lower is better). Rendering Performance averages the full $3\times3$ grid; Granularity Gains measure $P_0 \to P_2$ changes under $A_N$; Conflict Robustness is evaluated at $P_1$ under $A_C$. Overall weights the three perspectives by $0.50/0.25/0.25$ (details in \Cref{sec:appendix_evaluation}).}
\label{tab:main_results}
\vspace{-0.9em}
\scriptsize
\resizebox{\textwidth}{!}{%
\begin{tabular}{lc cccc ccc cccc c}
\hlineB{2.5}
\multirow{2}{*}{\textbf{Model}} & \multirow{2}{*}{\textbf{Size}}
& \multicolumn{4}{c}{\textbf{Rendering Performance}}
& \multicolumn{3}{c}{\textbf{Granularity Gains}}
& \multicolumn{4}{c}{\textbf{Conflict Robustness}}
& \multirow{2}{*}{\textbf{Overall}$\uparrow$} \\
\cmidrule(lr){3-6}\cmidrule(lr){7-9}\cmidrule(lr){10-13}
& & \textbf{Preservation}$\uparrow$ & \textbf{Realization}$\uparrow$ & \textbf{Quality}$\uparrow$ & \textbf{Persistence}$\uparrow$ & \textbf{Structure}$\uparrow$ & \textbf{Appearance}$\uparrow$ & \textbf{Binding}$\uparrow$ & \textbf{Adherence}$\uparrow$ & \textbf{Retention}$\uparrow$ & \textbf{Penalty}$\downarrow$ & \textbf{Leakage}$\downarrow$ & \\
\hlineB{1.5}
\multicolumn{14}{c}{\cellcolor{lightgray!20}\textit{Open-Weight Models}} \\
\omniweavinglogo~OmniWeaving & 8.3B & 61.61 & 66.37 & 62.02 & 58.01 & +12.75 & -14.47 & -15.95 & 58.13 & 47.68 & 10.63 & 50.97 & 57.62 \\
\univideologo~UniVideo & 13B & 53.34 & 67.31 & 63.01 & 59.10 & +15.61 & -4.24 & -7.43 & 63.39 & 51.78 & 6.02 & 52.24 & 57.82 \\
\vinologo~VINO & 13B & 52.83 & 67.95 & 63.43 & 60.38 & +15.99 & -2.70 & -5.21 & 67.34 & 56.75 & 3.57 & 38.26 & 59.78 \\
\wanlogo~Wan2.1-VACE & 14B & 60.97 & 60.86 & 60.78 & 51.68 & +18.82 & -2.89 & -2.68 & 60.93 & 50.37 & 8.05 & 64.17 & 56.89 \\
\wanlogo~Wan2.2-VACE & 14B & 60.52 & 70.25 & 64.11 & 62.09 & +11.48 & -2.37 & -4.74 & 68.80 & 59.18 & 3.14 & 33.60 & 62.43 \\
\ctorlogo~C2R & 14B & 53.05 & 71.14 & 65.14 & 64.96 & +15.98 & -2.75 & -5.65 & 69.95 & 63.11 & 4.45 & 43.03 & 61.10 \\
\berninilogo~Bernini-R & 14B & \hlsecond{61.81} & 72.41 & 63.88 & 64.80 & +8.45 & -3.33 & -3.51 & 71.17 & 62.53 & 2.93 & 43.16 & 62.97 \\
\lynnreallogo~LynnReal-Omni & 32B & 60.52 & 75.51 & \hlsecond{65.48} & 68.41 & +7.26 & -3.62 & -5.28 & 75.32 & 68.05 & 2.26 & 30.54 & \hlfirst{64.89} \\
\minimaxlogo~MiniMax-H3 & 33B & 50.70 & \hlsecond{76.38} & 64.06 & \hlsecond{70.32} & +8.01 & -0.89 & -1.93 & 76.41 & \hlsecond{70.80} & \hlsecond{0.74} & \hlsecond{28.64} & 63.63 \\
\solarlogo~SolarRenderer & 33B & \hlfirst{63.28} & 69.97 & 62.48 & 61.68 & +12.25 & -1.49 & -2.02 & 69.43 & 60.17 & 4.27 & 40.37 & 62.59 \\
\hline
\multicolumn{14}{c}{\cellcolor{lightgray!20}\textit{Commercial Models}} \\
\wanlogo~Wan~3.0 & - & 44.37 & \hlfirst{80.40} & \hlfirst{66.92} & \hlfirst{75.41} & -0.35 & -0.58 & -0.16 & \hlfirst{79.82} & \hlfirst{74.18} & \hlfirst{0.53} & \hlfirst{27.39} & 63.36 \\
\seedancelogo~Seedance~2.5 & - & 58.66 & 75.55 & 63.10 & 69.20 & +10.88 & +2.51 & +3.21 & \hlsecond{76.60} & 69.53 & 1.97 & 41.57 & \hlsecond{64.73} \\
\hlineB{2.5}
\end{tabular}%
}
\vspace{-0.8em}
\end{table*}
\endgroup

\vspace{-0.8em}
\paragraph{Evaluations.} Following \Cref{sec:evaluation_protocol}, we report aggregated dimensional metrics across three core perspectives:
\ding{182}~\textbf{Rendering Performance} summarizes overall competence across the full $3 \times 3$ grid via World Preservation ($\mathrm{WP}$: Entity, Motion, Camera), Rendering Fidelity ($\mathrm{RF}$), Visual Quality ($\mathrm{VQ}$), and Appearance Persistence ($\mathrm{AP}$);
\ding{183}~\textbf{Granularity Gains} measures endpoint gains ($\Delta_{02}$) across draft levels ($P_0 \to P_1 \to P_2$) under neutral appearance $A_N$; and
\ding{184}~\textbf{Conflict Robustness} assesses resilience to conflicting source cues $A_C$ at $P_1$ via $\mathrm{RF}_C$, $\mathrm{AP}_C$, Conflict Penalty ($\mathrm{CP}$), and Source Leakage ($\mathrm{SL}$).
All scores are macro-averaged within each family before aggregating across families on common cohorts. More details in \Cref{sec:appendix_evaluation}.
% See \textcolor{red}{Appendix} for detailed formulations.

\subsection{Main Results}

To systematically characterize the capability boundaries of video generative renderers, we first benchmark 12 video models across the $3 \times 3$ grid of \ourbench. Key results in \Cref{tab:main_results} and \Cref{fig:overall} yield three primary takeaways:

\vspace{-0.8em}
\paragraph{Takeaway \ding{182}: World preservation and visual realization remain in stark tension.}
Models face an inherent dilemma between retaining 4D physics and realizing target appearance.
As shown in the \textit{Rendering Performance} columns of \Cref{tab:main_results} and \Cref{fig:overall}, dedicated renderers (\eg, \solarlogo~SolarRenderer) lead in Preservation ($\text{WP} = 63.28$) but lag in Realization ($\text{RF} = 69.97$). Conversely, large models (\eg, \wanlogo~Wan~3.0) maximize Realization ($\text{RF} = 80.40$) at the cost of severe physical corruption ($\text{WP} = 44.37$, lowest overall).
Remarkably, even prevailing commercial leaders like \seedancelogo~Seedance~2.5, which dominate conventional video benchmarks, fail to top \ourbench (Overall: $64.73$, edged out by 32B \lynnreallogo~LynnReal-Omni at $64.89$), compromised by sub-optimal physical preservation ($\text{WP} = 58.66$). This confirms that generic visual scaling does not resolve the dual-correctness challenge.

\vspace{-0.8em}
\paragraph{Takeaway \ding{183}: Richer draft scaffolding anchors structure but impairs appearance binding.}
Advancing draft granularity ($P_0 \to P_2$ under $A_N$) reliably improves structure but yields negative appearance returns.
As reported in the \textit{Granularity Gains} of \Cref{tab:main_results} and \Cref{fig:overall}, structural gains ($g_s$) are consistently positive across open-weight models (\eg, $+18.82$ for \wanlogo~Wan2.1-VACE), confirming that dense meshes suppress physical drift.
However, appearance ($g_a$) and binding ($g_b$) gains are negative for 9 out of 10 models (\eg, \omniweavinglogo~OmniWeaving drops by $-14.47\,/\,-15.95$), with \seedancelogo~Seedance~2.5 as the sole exception ($+2.51\,/\,+3.21$). Denser drafts introduce synthetic shading biases that actively override text-conditioned cross-attention.

\begin{figure*}[!t]
\centering
% \vspace{-0.4em}
\includegraphics[width=\linewidth]{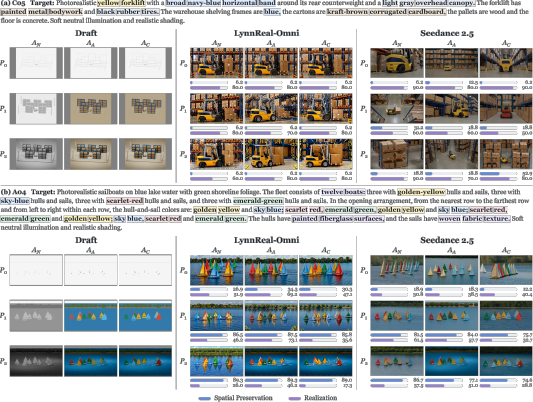}
\vspace{-2em}
% \vspace{-0.8em}
\caption{Failure cases demonstration of \lynnreallogo~LynnReal-Omni and \seedancelogo~Seedance~2.5 across the $3\times3$ grid on: (\textbf{\textit{a}}) C05 (\textit{Occlusion \& Reappearance}, warehouse forklift) and (\textbf{\textit{b}}) A04 (\textit{Multi-Entity Orchestration}, 12-sailboat fleet). More case studies are in \url{https://haroldchen19.github.io/WorldDraft-site}.}
\label{fig:failure}
\vspace{-0.8em}
\end{figure*}

\vspace{-0.8em}
\paragraph{Takeaway \ding{184}: Conflicting source cues trigger widespread leakage, exposing fragile disentanglement.}
Under contradictory source-target pairs ($A_C$ at $P_1$), models struggle to separate motion scaffolds from surface appearances.
As evidenced by the \textit{Conflict Robustness} metrics in \Cref{tab:main_results} and \Cref{fig:overall}, open-weight models suffer large Conflict Penalties (up to $10.63$ for \omniweavinglogo~OmniWeaving) and rampant color leakage (\eg, $64.17\%$ for \wanlogo~Wan2.1-VACE and $52.24\%$ for \univideologo~UniVideo). Even top commercial and large models (\eg, \wanlogo~Wan~3.0, \minimaxlogo~MiniMax-H3) exhibit $\sim 28\%\text{-}30\%$ leakage, revealing that renderers treat draft pixels as rigid visual constraints rather than overwritable scaffolds.

\subsection{Failure Mode Analysis}

To diagnose why state-of-the-art models struggle with generative rendering, \Cref{fig:failure} visualizes representative failure cases from top-performing models (\lynnreallogo~LynnReal-Omni and \seedancelogo~Seedance~2.5) on C05 (\textit{Occlusion \& Reappearance}) and A04 (\textit{Multi-Entity Orchestration}):

\vspace{-0.8em}
\paragraph{Failure Mode \ding{182}: Prompt-driven structural override.}
When processing dynamic scenes with complex layouts, models frequently sacrifice draft kinematics to satisfy textual semantics.
As illustrated in \Cref{fig:failure} (\textit{a}), both \lynnreallogo~LynnReal-Omni and \seedancelogo~Seedance~2.5 exhibit near-zero Spatial Preservation (consistently scoring $\le 18.8$, dipping to $6.2$).
Instead of adhering to the prescribed camera trajectory and warehouse shelving geometry, the models synthesize entirely ungrounded viewpoints and forklift trajectories conditioned solely on the prompt. This demonstrates that models easily devolve from faithful renderers into unconstrained text-to-video generators.

\vspace{-0.8em}
\paragraph{Failure Mode \ding{183}: Deficient spatial reasoning and attribute binding.} Under neutral scaffolds ($A_N$), models exhibit weak spatial understanding, failing to parse relative coordinate references and topological arrangements.
As shown in \Cref{fig:failure} (\textit{b}) , the prompt specifies an ordered $3 \times 4$ color layout across 12 sailboats.
Lacking the ability to resolve relative spatial hierarchies (\eg, row depths and left-to-right ordering), the models scramble target colors, causing severe attribute cross-contamination.

\vspace{-0.8em}
\paragraph{Failure Mode \ding{184}: Previs texture latching and conflicting leakage.}
Richer drafts and conflicting cues severely disrupt appearance synthesis.
In \Cref{fig:failure} (\textit{b}), advancing from simulation ($P_1$) to previs ($P_2$) causes Realization scores to collapse for both models (\eg, \lynnreallogo~LynnReal-Omni drops from $73.1$ to $46.2$ under $A_A$, and plummets to $17.3$ under $A_C$).
Video diffusion models over-attend to the dense synthetic shading and specular highlights of $P_2$ drafts, treating them as immutable visual priors.
Under conflict ($A_C$), models either persistently reproduce conflicting source colors (source leakage) or hallucinate unprescribed objects (\eg, \lynnreallogo~LynnReal-Omni synthesizing an extra cluster of green sailboats at $P_1, A_C$) to compensate for text prompts, directly violating draft topology.

\begingroup
\setlength{\tabcolsep}{1.6pt}
\begin{table*}[t]
% \vspace{-0.6em}
\renewcommand{\arraystretch}{1.6}
\centering
\caption{Evaluation on \ourbench with \ourmethod. We \textbf{bold} the best results. Qualitative results are in \Cref{fig:primer_1,fig:primer_2}.}
\label{tab:primer}
\vspace{-0.9em}
\scriptsize
\resizebox{\textwidth}{!}{%
\begin{tabular}{l cccc ccc cccc c}
\hlineB{2.5}
\multirow{2}{*}{\textbf{Model}} & \multicolumn{4}{c}{\textbf{Rendering Performance}}
& \multicolumn{3}{c}{\textbf{Granularity Gains}}
& \multicolumn{4}{c}{\textbf{Conflict Robustness}}
& \multirow{2}{*}{\textbf{Overall}$\uparrow$} \\
\cmidrule(lr){2-5}\cmidrule(lr){6-8}\cmidrule(lr){9-12}
& \textbf{Preservation}$\uparrow$ & \textbf{Realization}$\uparrow$ & \textbf{Quality}$\uparrow$ & \textbf{Persistence}$\uparrow$ & \textbf{Structure}$\uparrow$ & \textbf{Appearance}$\uparrow$ & \textbf{Binding}$\uparrow$ & \textbf{Adherence}$\uparrow$ & \textbf{Retention}$\uparrow$ & \textbf{Penalty}$\downarrow$ & \textbf{Leakage}$\downarrow$ & \\
\hlineB{1.5}
\lynnreallogo~LynnReal-Omni & 60.52 & 75.51 & 65.48 & 68.41 & +7.26 & -3.62 & -5.28 & 75.32 & 68.05 & 2.26 & 30.54 & 64.89 \\
\seedancelogo~Seedance~2.5 & 58.66 & 75.55 & 63.10 & 69.20 & +10.88 & +2.51 & +3.21 & 76.60 & 69.53 & 1.97 & 41.57 & 64.73 \\
\hline
\omniweavinglogo~OmniWeaving & 61.61 & 66.37 & 62.02 & 58.01 & +12.75 & -14.47 & -15.95 & 58.13 & 47.68 & 10.63 & 50.97 & 57.62 \\
$+$ Text PE & 62.63 & 67.98 & 63.46 & 59.36 & +12.98 & -14.81 & -16.32 & 60.06 & 48.79 & \textbf{10.30} & 49.79 & 58.52 \\
\rowcolor{cyan!10}
$+$ \ourmethod & \textbf{64.60} & \textbf{70.93} & \textbf{66.19} & \textbf{61.90} & +13.51 & -15.45 & -17.02 & \textbf{62.97} & \textbf{50.88} & 10.41 & \textbf{47.55} & \textbf{60.13} \\
\hline
\minimaxlogo~MiniMax-H3 & 50.70 & 76.38 & 64.06 & 70.32 & +8.01 & -0.89 & -1.93 & 76.41 & 70.80 & \textbf{0.74} & 28.64 & 63.63 \\
$+$ Text PE & 51.45 & 77.91 & 65.34 & 71.72 & +8.07 & -0.91 & -1.97 & 77.94 & 72.22 & 0.75 & 28.07 & 64.89 \\
\rowcolor{cyan!10}
$+$ \ourmethod & \textbf{52.55} & \textbf{80.21} & \textbf{67.26} & \textbf{73.84} & +8.24 & -0.93 & -2.02 & \textbf{80.23} & \textbf{74.34} & 0.78 & \textbf{27.21} & \textbf{65.56} \\
\hlineB{2.5}
\end{tabular}%
}
\vspace{-0.8em}
\end{table*}
\endgroup

\subsection{Results with \ourmethod}

Motivated by the failure modes above, we further sought to investigate whether a lightweight test-time strategy can unlock frozen video renderers without retraining.
To this end, we evaluate \ourmethod across representative models (\omniweavinglogo~OmniWeaving and \minimaxlogo~MiniMax-H3) against raw baselines and text prompt engineering (Text PE); results are in \Cref{tab:primer} and \Cref{fig:primer_1,fig:primer_2}.

\vspace{-0.8em}
\paragraph{\ourmethod consistently enhances both world preservation and visual realization.} As shown in \Cref{tab:primer}, \ourmethod consistently boosts both world preservation and appearance realization across architectures.
For \omniweavinglogo~OmniWeaving, \ourmethod enhances Preservation ($+2.99$) and Realization ($+4.56$) while reducing Source Leakage by $3.42\%$, yielding an Overall gain of $+2.51$.
For \minimaxlogo~MiniMax-H3, it boosts Realization to $80.21$ ($+3.83$) and elevates the Overall score to $65.56$ ($+1.93$), surpassing both \lynnreallogo~LynnReal-Omni ($64.89$) and the commercial leader \seedancelogo~Seedance~2.5 ($64.73$).
Conversely, Text PE provides negligible improvements ($+0.90$ and $+1.26$), limited by the inherent spatial ambiguity of text.
Qualitative results in \Cref{fig:primer_1,fig:primer_2} further confirm these findings: first-frame cues effectively stabilize global camera framing against unprompted zoom-ins, prevent multi-entity attribute cross-contamination (\eg, binding cobalt-blue exclusively to the rightmost ball in B02), and suppress deceptive source color leakage under $A_C$.

% \subsection{Further Analysis}

\section{Conclusion}

In this work, we explored the emerging division of labor between code-driven simulation and video generation by proposing \ourbench, a systematic benchmark of $432$ strictly paired test cases across $48$ executable 4D world families.
Extensive evaluations across $12$ representative models yield clear answers to our three foundational questions:
\ding{192}~world preservation and target appearance realization remain in stark operational tension;
\ding{193}~richer draft scaffolding reliably anchors structure but paradoxically impairs appearance fidelity; and
\ding{194}~conflicting source cues trigger severe source leakage, exposing fragile cross-modal disentanglement.
To mitigate these bottlenecks, we introduce \ourmethod, a plug-and-play visual prompting strategy that consistently improves both objectives across frozen video renderers.
Together, \ourbench and \ourmethod establish a foundational stepping stone toward physically grounded generative rendering.

% \clearpage
\newpage
\bibliographystyle{assets/plainnat}
\bibliography{paper}

@String(CVPR= {IEEE Conf. Comput. Vis. Pattern Recog.})

@String(ICCV= {Int. Conf. Comput. Vis.})

@String(AAAI = {AAAI})

@String(CVPR  = {CVPR})

@String(ICCV  = {ICCV})

@article{huang2026rowbench,
  title={ROWBench: Do Video Models Render What the Program Specifies?},
  author={Huang, Zheng-Hui and Lin, Guixu and Tsai, Yu-Ju and Zhu, Jian-Kai and Lan, Fengbo and Liu, Yu-Lun and Chuang, Yung-Yu and Zhang, Kaipeng and Wang, Zhixiang},
  journal={arXiv preprint arXiv:2610.02205},
  year={2026}
}

@inproceedings{cheng2025vpo,
  title={Vpo: Aligning text-to-video generation models with prompt optimization},
  author={Cheng, Jiale and Lyu, Ruiliang and Gu, Xiaotao and Liu, Xiao and Xu, Jiazheng and Lu, Yida and Teng, Jiayan and Yang, Zhuoyi and Dong, Yuxiao and Tang, Jie and others},
  booktitle={2025 IEEE/CVF International Conference on Computer Vision (ICCV)},
  pages={15636--15645},
  year={2025},
  organization={IEEE}
}

@article{schuhmann2022laion,
  title={Laion-5b: An open large-scale dataset for training next generation image-text models},
  author={Schuhmann, Christoph and Beaumont, Romain and Vencu, Richard and Gordon, Cade and Wightman, Ross and Cherti, Mehdi and Coombes, Theo and Katta, Aarush and Mullis, Clayton and Wortsman, Mitchell and others},
  journal={Advances in neural information processing systems},
  volume={35},
  pages={25278--25294},
  year={2022}
}

@inproceedings{ke2021musiq,
  title={Musiq: Multi-scale image quality transformer},
  author={Ke, Junjie and Wang, Qifei and Wang, Yilin and Milanfar, Peyman and Yang, Feng},
  booktitle={2021 IEEE/CVF International Conference on Computer Vision (ICCV)},
  pages={5128--5137},
  year={2021},
  organization={IEEE}
}

@inproceedings{huang2024vbench,
  title={Vbench: Comprehensive benchmark suite for video generative models},
  author={Huang, Ziqi and He, Yinan and Yu, Jiashuo and Zhang, Fan and Si, Chenyang and Jiang, Yuming and Zhang, Yuanhan and Wu, Tianxing and Jin, Qingyang and Chanpaisit, Nattapol and others},
  booktitle={2024 IEEE/CVF Conference on Computer Vision and Pattern Recognition (CVPR)},
  pages={21807--21818},
  year={2024},
  organization={IEEE}
}

@article{huang2025vipe,
  title={Vipe: Video pose engine for 3d geometric perception},
  author={Huang, Jiahui and Zhou, Qunjie and Rabeti, Hesam and Korovko, Aleksandr and Ling, Huan and Ren, Xuanchi and Shen, Tianchang and Gao, Jun and Slepichev, Dmitry and Lin, Chen-Hsuan and others},
  journal={arXiv preprint arXiv:2508.10934},
  year={2025}
}

@inproceedings{carion2026sam,
  title={Sam 3: Segment anything with concepts},
  author={Carion, Nicolas and Gustafson, Laura and Hu, Yuan-Ting and Debnath, Shoubhik and Hu, Ronghang and Suris Coll-Vinent, Didac and Ryali, Chaitanya and Alwala, Kalyan Vasudev and Khedr, Haitham and Huang, Andrew and others},
  booktitle={International conference on learning representations},
  volume={2026},
  pages={138846--138923},
  year={2026}
}

@misc{bytedance2025seedance25,
  author       = {{ByteDance Seed}},
  title        = {Seedance 2.5},
  year         = {2025},
  howpublished = {\url{https://seed.bytedance.com/en/seedance2_5}},
  note         = {Accessed: 2026-11-03}
}

@inproceedings{jiang2025vace,
  title={Vace: All-in-one video creation and editing},
  author={Jiang, Zeyinzi and Han, Zhen and Mao, Chaojie and Zhang, Jingfeng and Pan, Yulin and Liu, Yu},
  booktitle={2025 IEEE/CVF International Conference on Computer Vision (ICCV)},
  pages={17191--17202},
  year={2025},
  organization={IEEE}
}

@article{team2026bernini,
  title={Bernini: Latent Semantic Planning for Video Diffusion},
  author={Team, Bernini and Liu, Chenchen and Chen, Junyi and Li, Lei and Chi, Lu and Sun, Mingzhen and Li, Zhuoying and Fu, Yi and Guo, Ruoyu and Wu, Yiheng and others},
  journal={arXiv preprint arXiv:2605.22344},
  year={2026}
}

@inproceedings{wei2026univideo,
  title={Univideo: Unified understanding, generation, and editing for videos},
  author={Wei, Cong and Liu, Quande and Ye, Zixuan and Wang, Qiulin and Wang, Xintao and Wan, Pengfei and Gai, Kun and Chen, Wenhu},
  booktitle={International Conference on Learning Representations},
  volume={2026},
  pages={113905--113933},
  year={2026}
}

@article{chen2026vino,
  title={Vino: A unified visual generator with interleaved omnimodal context},
  author={Chen, Junyi and He, Tong and Fu, Zhoujie and Wan, Pengfei and Gai, Kun and Ye, Weicai},
  journal={arXiv preprint arXiv:2601.02358},
  year={2026}
}

@article{mao2026lynnreal,
  title={LynnReal-Omni: Native multi-modal Video Generation for Agentic Visual Workflows},
  author={Mao, Xiaofeng and Lin, Peijia and Rui, Shaohao and Zhang, Yibo and Wan, Haibin and Ma, Weijie},
  journal={arXiv preprint arXiv:2609.15863},
  year={2026}
}

@article{fang2026solarrenderer,
  title={SolarRenderer: Few-Step Generative Reference Rendering},
  author={Fang, Guian and Qian, Shengju and Lu, Dongyue and Huang, Junchao and Jiang, Lutao and Wang, Xin and Song, Yiren and Jiang, Li and Shou, Mike Zheng},
  journal={arXiv preprint arXiv:2610.28630},
  year={2026}
}

@misc{minimax2025minimaxh3,
  author       = {{MiniMax}},
  title        = {MiniMax-H3},
  year         = {2025},
  howpublished = {\url{https://www.minimax.io/blog/minimax-h3}},
  note         = {Accessed: 2026-11-03}
}

@article{yang2025qwen3,
  title={Qwen3 technical report},
  author={Yang, An and Li, Anfeng and Yang, Baosong and Zhang, Beichen and Hui, Binyuan and Zheng, Bo and Yu, Bowen and Gao, Chang and Huang, Chengen and Lv, Chenxu and others},
  journal={arXiv preprint arXiv:2505.09388},
  year={2025}
}

@article{qi2026mme,
  title={MME-CoF-Pro: Evaluating Reasoning Coherence in Video Generative Models with Text and Visual Hints},
  author={Qi, Yu and Xu, Xinyi and Guo, Ziyu and Ma, Siyuan and Zhang, Renrui and Chen, Xinyan and An, Ruichuan and Xing, Ruofan and Zhang, Jiayi and Huang, Haojie and others},
  journal={arXiv preprint arXiv:2603.20194},
  year={2026}
}

@article{fang2025video,
  title={In-Video Instructions: Visual Signals as Generative Control},
  author={Fang, Gongfan and Ma, Xinyin and Wang, Xinchao},
  journal={arXiv preprint arXiv:2511.19401},
  year={2025}
}

@article{zhang2026gs,
  title={GS-Agent: Creating 4D Physical Worlds With Generative Simulation},
  author={Zhang, Hongxin and Lin, Chunru and Li, Junyan and Xian, Zhou and Wang, Tsun-Hsuan and Gan, Chuang},
  journal={arXiv preprint arXiv:2607.21522},
  year={2026}
}

@inproceedings{radford2021learning,
  title={Learning transferable visual models from natural language supervision},
  author={Radford, Alec and Kim, Jong Wook and Hallacy, Chris and Ramesh, Aditya and Goh, Gabriel and Agarwal, Sandhini and Sastry, Girish and Askell, Amanda and Mishkin, Pamela and Clark, Jack and others},
  booktitle={International conference on machine learning},
  pages={8748--8763},
  year={2021},
  organization={PmLR}
}

@article{unterthiner2018towards,
  title={Towards accurate generative models of video: A new metric \& challenges},
  author={Unterthiner, Thomas and Van Steenkiste, Sjoerd and Kurach, Karol and Marinier, Raphael and Michalski, Marcin and Gelly, Sylvain},
  journal={arXiv preprint arXiv:1812.01717},
  year={2018}
}

@article{gao2026vefx,
  title={VEFX-Bench: A Holistic Benchmark for Generic Video Editing and Visual Effects},
  author={Gao, Xiangbo and Jiang, Sicong and Liu, Bangya and Chen, Xinghao and Yang, Minglai and Yang, Siyuan and Wu, Mingyang and Yu, Jiongze and Zheng, Qi and Wang, Haozhi and others},
  journal={arXiv preprint arXiv:2604.16272},
  year={2026}
}

@inproceedings{chen2025editboard,
  title={Editboard: Towards a comprehensive evaluation benchmark for text-based video editing models},
  author={Chen, Yupeng and Chen, Penglin and Zhang, Xiaoyu and Huang, Yixian and Xie, Qian},
  booktitle={Proceedings of the AAAI Conference on Artificial Intelligence},
  volume={39},
  number={15},
  pages={15975--15983},
  year={2025}
}

@article{wang20263d,
  title={3D-CoS: A New 3D Reconstruction Paradigm Based on VLM Code Synthesis},
  author={Wang, Yuhao and Wang, Puyi and Li, Linjie and Yang, Zhengyuan and Lin, Kevin Qinghong and Cheng, Yu},
  journal={arXiv preprint arXiv:2606.10478},
  year={2026}
}

@inproceedings{huang2024blenderalchemy,
  title={Blenderalchemy: Editing 3d graphics with vision-language models},
  author={Huang, Ian and Yang, Guandao and Guibas, Leonidas},
  booktitle={European Conference on Computer Vision},
  pages={297--314},
  year={2024},
  organization={Springer}
}

@article{wan2025wan,
  title={Wan: Open and advanced large-scale video generative models},
  author={Wan, Team and Wang, Ang and Ai, Baole and Wen, Bin and Mao, Chaojie and Xie, Chen-Wei and Chen, Di and Yu, Feiwu and Zhao, Haiming and Yang, Jianxiao and others},
  journal={arXiv preprint arXiv:2503.20314},
  year={2025}
}

@article{hacohen2024ltx,
  title={Ltx-video: Realtime video latent diffusion},
  author={HaCohen, Yoav and Chiprut, Nisan and Brazowski, Benny and Shalem, Daniel and Moshe, Dudu and Richardson, Eitan and Levin, Eran and Shiran, Guy and Zabari, Nir and Gordon, Ori and others},
  journal={arXiv preprint arXiv:2501.00103},
  year={2024}
}

@article{pan2026omniweaving,
  title={Omniweaving: Towards unified video generation with free-form composition and reasoning},
  author={Pan, Kaihang and Tian, Qi and Zhang, Jianwei and Kong, Weijie and Xiong, Jiangfeng and Long, Yanxin and Zhang, Shixue and Qiu, Haiyi and Wang, Tan and Lv, Zheqi and others},
  journal={arXiv preprint arXiv:2603.24458},
  year={2026}
}

@article{liu2026v2v,
  title={V2V-Bench: A Comprehensive Benchmark for Video-to-Video Generation Evaluation},
  author={Liu, Tao and Krishna, Leela and Kumar, Gouti Pavan and Garg, Vishav and others},
  journal={arXiv preprint arXiv:2606.05665},
  year={2026}
}

@article{gomez2026coarse,
  title={Coarse-to-Real: Generative Rendering for Populated Dynamic Scenes},
  author={Gomez-Nogales, Gonzalo and Hong, Yicong and Ge, Chongjian and Zhuang, Peiye and Comino-Trinidad, Marc and Casas, Dan and Zhou, Yi},
  journal={arXiv preprint arXiv:2601.22301},
  year={2026}
}

@inproceedings{li2025five,
  title={Five-bench: A fine-grained video editing benchmark for evaluating emerging diffusion and rectified flow models},
  author={Li, Minghan and Xie, Chenxi and Wu, Yichen and Zhang, Lei and Wang, Mengyu},
  booktitle={2025 IEEE/CVF International Conference on Computer Vision (ICCV)},
  pages={16672--16681},
  year={2025},
  organization={IEEE}
}

@article{singh2025openai,
  title={Openai gpt-5 system card},
  author={Singh, Aaditya and Fry, Adam and Perelman, Adam and Tart, Adam and Ganesh, Adi and El-Kishky, Ahmed and McLaughlin, Aidan and Low, Aiden and Ostrow, AJ and Ananthram, Akhila and others},
  journal={arXiv preprint arXiv:2601.03267},
  year={2025}
}

@misc{liang2026codevideomodel,
  title        = {Code Video Model},
  author       = {Liang, Zhiyang and Zhang, Jiawei and Qian, Guocheng and Chen, Dongdong and Wan, Ziyu},
  year         = {2026},
  month        = oct,
  note         = {Preprint},
  doi          = {10.13140/RG.2.2.34014.80961},
  url          = {https://code-video-model.github.io/tech_report.pdf}
}

@article{li2026videococo,
  title={Videococo: Code-as-cot for physically-consistent video generation via an agentic dual-engine system},
  author={Li, Haodong and Ren, Tianfei and Ma, Xiaoxiao and Qing, Chunmei and Fang, Zhen and He, Sipeng and Guo, Ziyu and Wu, Haoyu and Tian, Juanxi and Zou, Yihang and others},
  journal={arXiv preprint arXiv:2607.27380},
  year={2026}
}

@article{wang2026code,
  title={Code as Worlds: Agentic discovery of executable world representations for physical reasoning},
  author={Wang, Hanyang and Cai, Yimo and Chen, Weiliang and Chi, Jiawei and Sun, Haowen and Dai, Qiyu and Hung, Yi-Hsin and Guo, Xingzhuo and Ren, Jinshan and Yao, Runmao and others},
  journal={arXiv preprint arXiv:2608.27549},
  year={2026}
}

@inproceedings{yin2026vision,
  title={Vision-as-inverse-graphics agent via interleaved multimodal reasoning},
  author={Yin, Shaofeng and Ge, Jiaxin and Wang, Zora Zhiruo and Wang, Chenyang and Li, Xiuyu and Black, Michael J and Darrell, Trevor and Kanazawa, Angjoo and Feng, Haiwen},
  booktitle={European Conference on Computer Vision},
  pages={557--575},
  year={2026},
  organization={Springer}
}

@article{
li2026video,
title={Video Generation Models: A Survey of Post-Training and Alignment},
author={Chaoyu Li and Xiaoyi Gu and Yogesh Kulkarni and Eun Woo Im and Mohammadmahdi Honarmand and Zeyu Wang and Juntong Song and Fei Du and Xilin Jiang and Kexin Zheng and Tianzhi Li and Fei Tao and Pooyan Fazli},
journal={Transactions on Machine Learning Research},
issn={2835-8856},
year={2026},
url={https://openreview.net/forum?id=YlUEWLESIu},
note={}
}

@article{agarwal2025cosmos,
  title={Cosmos world foundation model platform for physical ai},
  author={Agarwal, Niket and Ali, Arslan and Bala, Maciej and Balaji, Yogesh and Barker, Erik and Cai, Tiffany and Chattopadhyay, Prithvijit and Chen, Yongxin and Cui, Yin and Ding, Yifan and others},
  journal={arXiv preprint arXiv:2501.03575},
  year={2025}
}

@inproceedings{todorov2012mujoco,
  title={Mujoco: A physics engine for model-based control},
  author={Todorov, Emanuel and Erez, Tom and Tassa, Yuval},
  booktitle={2012 IEEE/RSJ international conference on intelligent robots and systems},
  pages={5026--5033},
  year={2012},
  organization={IEEE}
}

@misc{BlenderOnlineCommunity2026,
  author = {{Blender Online Community}},
  title = {Blender -- a 3D Modelling and Rendering Package},
  year = {2026},
  publisher = {Blender Foundation},
  address = {Stichting Blender Foundation, Amsterdam},
  url = {https://www.blender.org}
}

\clearpage
\appendix
% \setcounter{page}{1}
% \maketitlesupplementary

\section{More Details of \ourbench}
\label{sec:appendix_case_demo}

\subsection{Motivation of Each Challenge}

Grounded in the dual objectives of \textit{world preservation} and \textit{visual realization}, \ourbench curates dynamic scenarios that jointly impose deterministic physical constraints and multi-modal binding stressors. 
Evaluating generative renderers on these scenarios probes spatiotemporal reliability across diverse physical regimes: capabilities that conventional static image metrics fundamentally fail to measure.
Below, we detail the design motivations behind our six dynamic challenge categories:

\vspace{-0.4em}
\begin{coveragebox}[title={I. Multi-Entity Orchestration}]
\noindent
When scenes contain multiple concurrent entities, particularly those with identical geometries, dense clusters, or dynamic spatial swaps, models face severe risks of identity confusion and attribute cross-contamination. 
This challenge explicitly assesses whether a renderer preserves exact instance counts, relative spatial layouts, and coordinated multi-agent trajectories, while accurately binding distinct target attributes (\eg, individualized colors and materials) to their designated semantic roles without spatial bleeding or entity dropping.
\end{coveragebox}
% \vspace{-0.2em}

\vspace{-0.6em}
\begin{coveragebox}[title={II. Multi-Stage Dynamics}]
\noindent
Locally smooth motion does not imply global causal correctness: models often generate visually plausible micro-movements while skipping discrete phase transitions or reversing event sequences.
This category evaluates whether key temporal boundaries, causal action-reaction orders, and phase-dependent kinematics are faithfully preserved, while simultaneously requiring target appearances to remain invariant across state-changing boundaries (\eg, a mechanical latch unlocking, opening, and locking).
\end{coveragebox}

\vspace{-0.6em}
\begin{coveragebox}[title={III. Occlusion \& Reappearance}]
\noindent
Physical occlusion temporarily removes direct visual evidence of an entity, breaking adjacent-frame continuity and demanding long-range spatiotemporal memory. 
This challenge measures whether the renderer maintains physical occlusion orders, follows emergence trajectories, and accurately re-identifies the target entity upon reappearance: restoring its precise target color, material, and surface texture rather than resetting or re-synthesizing a generic object.
\end{coveragebox}

\vspace{-0.6em}
\begin{coveragebox}[title={IV. Camera Motion}]
\noindent
Aggressive 6-DoF camera trajectories continuously reshape perspective projections, surface visibility, and parallax depth. 
This category evaluates whether the model faithfully preserves the canonical camera path and epipolar geometry. It demands strict multi-view appearance consistency, verifying that intrinsic material identities remain stable while view-dependent visual cues (\eg, specular highlights and Fresnel reflections) deform realistically under dynamic perspective shifts.
\end{coveragebox}

% \vspace{-0.8em}
\vspace{-0.6em}
\begin{coveragebox}[title={V. Non-Rigid Deformation}]
\noindent
Unlike rigid bodies with static topologies, soft-body kinematics continuously deform mesh geometries and local surface coordinates. 
This challenge tests whether continuous topological deformations and structural integrity are preserved, and whether target textures, patterns, and elastic materials conform naturally to the deforming substrate without visual slipping, surface tearing, or texture resetting.
\end{coveragebox}

% \vspace{-0.8em}
\vspace{-0.6em}
\begin{coveragebox}[title={VI. Hybrid Interactions}]
\noindent
Violent physical interactions, such as contact collisions, momentum transfer, and friction, couple the motions of disparate objects into codependent physical systems. Generating plausible individual motions in isolation easily violates joint Newtonian laws.
This category focuses on preserving contact boundaries, interaction outcomes, and kinetic energy exchange, while testing whether entity identities and materials survive intense physical contact. It directly complements Multi-Entity Orchestration by emphasizing dynamic force coupling over static spatial coordination.
\end{coveragebox}

\noindent Each executable world family in \ourbench is categorized by its primary physical stressor, while naturally incorporating secondary dynamics. \Cref{sec:appendix_case_demos} illustrates representative cases and paired draft variants.

\subsection{Data Demonstration}
\label{sec:appendix_case_demos}

% \Cref{fig:data_1,fig:data_2,fig:data_3,fig:data_4,fig:data_5,fig:data_6} 
\url{https://haroldchen19.github.io/WorldDraft-site} showcase representative cases across all six challenge categories, visualizing the real-world reference footage, target prompt, and the full $3 \times 3$ draft grid ($P_0 \to P_2 \times A_N, A_A, A_C$) for each world.

\section{More Details of Evaluation Suite}
\label{sec:appendix_evaluation}

This section details the formal definitions, score normalization, and aggregation protocols underlying the three evaluation perspectives and the composite score in \Cref{tab:main_results}.
Formally, let $c = (P, A) \in \Omega$ denote an input condition, where $\Omega = \{P_0, P_1, P_2\} \times \{A_N, A_A, A_C\}$ represents the full $3 \times 3$ grid of 9 paired configurations across $48$ executable world families $\{\mathcal{W}_i\}_{i=1}^{48}$. All rate- and accuracy-based metrics are reported on a standard $[0, 100]$ scale.

\vspace{-1em}
\paragraph{Shared Temporal \& Spatial Protocol.}
To ensure fair cross-model comparisons regardless of native frame rates, all evaluations operate on the first $5$ seconds of generated videos using absolute timestamp alignment rather than frame indices:
\begin{itemize}
    \item \textbf{Continuous Tracking \& Geometry:} Dense geometric evaluation (\eg, bounding boxes, tracks, and 6-DoF camera trajectories) is conducted at $8$\,Hz across $40$ uniformly spaced timestamps ($t \in [0, 4.875]$\,s).
    \item \textbf{Sparse Attribute \& Event QA:} Appearance and interaction probes evaluate $8$ key timestamps distributed across the $5$-second duration. Following \citep{huang2026rowbench}, question answering is executed via Qwen3.6-27B \citep{yang2025qwen3}. Visual inputs comprise both full frames and reference-grounded entity crops. To eliminate response bias, all probes incorporate strictly balanced affirmative and literal-negation questions.
    \item \textbf{Strict Pre-registration:} No output-dependent spatial alignment, post-hoc warping, or scale re-normalization is performed; models are strictly penalized for spatial drift, aspect distortion, or event desynchronization.
\end{itemize}
Below, we detail the metric composition and normalization for each perspective.

\subsection{Rendering Performance}
\label{sec:appendix_rendering}

The Rendering Performance section in \Cref{tab:main_results} evaluates foundational generation capability across four dimensional scores: \textit{Preservation} ($\mathrm{WP}$), \textit{Realization} ($\mathrm{RF}$), \textit{Quality} ($\mathrm{VQ}$), and \textit{Persistence} ($\mathrm{AP}$).

\vspace{-1em}
\paragraph{World Preservation.}
$\mathrm{WP}$ quantifies geometric, physical, and camera fidelity against simulator ground truth, hierarchically combining three sub-components:
\begin{equation}
\mathrm{WP}_c = 0.40 E_c + 0.40 M_c + 0.20 K_c.
\end{equation}

\noindent\ding{192} \textit{Entity and Layout ($E$):} We combine Global Count ($\mathrm{GC}$), Local Count ($\mathrm{LC}$), and Bounding Box IoU ($\mathrm{BI}$):
\begin{equation}
E_c = 0.25\,\mathrm{GC}_c + 0.25\,\mathrm{LC}_c + 0.50\,\mathrm{BI}_c.
\end{equation}
$\mathrm{GC}$ requires exact instance count agreement across all three sampled timestamps $\mathcal{T}_{\mathrm{cnt}} = \{0, 2.125, 5\}\,\text{s}$, while $\mathrm{LC}$ evaluates per-frame count accuracy within reference-defined spatial regions:
\begin{equation}
\begin{split}
\mathrm{GC} = 100\,\mathbb{E}_{q}\left[\prod_{t\in\mathcal{T}_{\mathrm{cnt}}}\mathbf{1}[\hat{n}_{q,t}=n_{q,t}]\right],\\
\mathrm{LC} = 100\,\mathbb{E}_{q,t}\left[\mathbf{1}[\hat{n}_{q,t}=n_{q,t}]\right].
\end{split}
\end{equation}
$\mathrm{BI}$ evaluates tracked 2D bounding boxes against simulator-projected amodal 3D entity bounds (including deformation). Entities are detected via SAM3 \citep{carion2026sam} using attribute-agnostic category prompts and matched across time using global Hungarian assignment (IoU threshold $\ge 0.05$):
\begin{equation}
\mathrm{BI} = 100\,\mathbb{E}_{e}\left[\frac{1}{|\mathcal{T}_e|}\sum_{t\in\mathcal{T}_e}\mathrm{IoU}(B_{e,t}, \hat{B}_{a(e),t})\right].
\end{equation}
Unmatched entities and missing boxes are scored as $0$. 
% When SAM3 fails to detect a category, SAM-2.1 is initialized from the first valid reference box as a tracking fallback.

\vspace{2pt}
\noindent\ding{193} \textit{Motion and Events ($M$):} We assess dynamic continuity and physical causality via Matched Track Coverage ($\mathrm{TC}$), Interaction Success Rate ($\mathrm{ISR}$), and Event Timing Coverage ($\mathrm{EC}$):
\begin{equation}
M_c = 0.25\,\mathrm{TC}_c + 0.50\,\mathrm{ISR}_c + 0.25\,\mathrm{EC}_c.
\end{equation}
$\mathrm{TC} = 100\,\mathbb{E}_{e\in\mathcal{O}_{\mathrm{dyn}}}[\frac{1}{|\mathcal{T}_e|}\sum_{t\in\mathcal{T}_e}\mathbf{1}[\hat{B}_{a(e),t} \text{ exists}]]$ measures the temporal presence of tracked dynamic entities under the accepted assignment. $\mathrm{ISR} = 100\,\mathbb{E}_{q}[\mathbf{1}[\hat{a}_q = a_q]\,\mathbf{1}[\hat{o}_q = o_q]]$ validates that both the dynamic action ($a_q$) and its physical outcome ($o_q$) are simultaneously correct over 12 uniform frames. $\mathrm{EC} = 100 \cdot (N_{\mathrm{localized}} / N_{\mathrm{expected}})$ reports the fraction of causal events whose temporal boundary intervals $[t_{\mathrm{start}}, t_{\mathrm{end}}]$ are successfully localized.

\vspace{2pt}
\noindent\ding{194} \textit{Camera Kinematics ($K$):} Evaluated across the 8 camera-motion families by estimating 6-DoF trajectories $(\mathbf{R}(t), \mathbf{t}(t))$ via ViPE \citep{huang2025vipe} against the canonical draft. Camera translation is scale-normalized by trajectory peak norm: $\mathbf{u}(t) = \mathbf{t}(t) / (\max_{\tau}\|\mathbf{t}(\tau)\|_2 + 10^{-9})$. The camera fidelity score $K \in [0, 100]$ maps rotation error $\mathrm{Rot} \in [0, 180^\circ]$ and normalized translation error $\mathrm{Trans} \in [0, 2]$ to:
\begin{equation}
K = 100\left[1 - \frac{1}{2}\left(\frac{\mathrm{Rot}}{180} + \frac{\mathrm{Trans}}{2}\right)\right].
\end{equation}

\vspace{-1em}
\paragraph{Rendering Fidelity.}
Rendering Fidelity evaluates compliance with target color, material, texture, environment, and lighting specifications via balanced binary VLM QA on source-blind judges:
\begin{equation}
\mathrm{RF} = \frac{100}{|\mathcal{Q}_{\mathrm{RF}}|}\sum_{q\in\mathcal{Q}_{\mathrm{RF}}}\mathbf{1}[\hat{y}_q = y_q].
\end{equation}
Each probe evaluates visual evidence at the timestamp where reference entity visibility is maximized.

\vspace{-1em}
\paragraph{Visual Quality.}
Following VBench~\citep{huang2024vbench}, we compute frame-level technical and aesthetic quality across all native frames $\mathcal{F}$ in the first 5 seconds using MUSIQ-SPAQ \citep{ke2021musiq} ($\mathrm{IQ} \in [0, 100]$) and LAION \citep{schuhmann2022laion} aesthetic predictor on CLIP ViT-L/14 \citep{radford2021learning} ($\mathrm{AQ} = 10 \times \mathrm{Aesthetic} \in [0, 100]$):
\begin{equation}
\mathrm{VQ} = \frac{1}{2}\left(\frac{1}{|\mathcal{F}|}\sum_{f\in\mathcal{F}}\mathrm{IQ}(f) + \frac{1}{|\mathcal{F}|}\sum_{f\in\mathcal{F}}\mathrm{AQ}(f)\right).
\end{equation}
% Multi-frame CLIP text-visual alignment is logged independently and excluded from $\mathrm{VQ}$ to prevent semantic confounding.

\vspace{-1em}
\paragraph{Appearance Persistence.}
$\mathrm{AP}$ measures whether a correctly instantiated target attribute persistently binds to the entity over time:
\begin{equation}
\mathrm{AP} = \frac{100}{|\mathcal{Q}_{\mathrm{AP}}|}\sum_{q\in\mathcal{Q}_{\mathrm{AP}}} c_q \cdot s_q,
\end{equation}
where $c_q \in \{0, 1\}$ verifies attribute realization at a visible anchor frame, and $s_q \in \{0, 1\}$ verifies temporal consistency across subsequent visible timestamps. An incorrect anchor ($c_q = 0$) strictly zeroes out the score.

\vspace{-1em}
\paragraph{Aggregation.}
For any metric $X \in \{\mathrm{WP}, \mathrm{RF}, \mathrm{VQ}, \mathrm{AP}\}$, we first macro-average per-video scores across its applicable world family cohort $\mathcal{I}_X$ for each condition $c \in \mathcal{C}$:
\begin{equation}
X_c = \frac{1}{|\mathcal{I}_X|}\sum_{i\in\mathcal{I}_X} s^{(X)}_{i, c}.
\end{equation}
The four Rendering Performance columns in \Cref{tab:main_results} report the mean score over the full $3 \times 3$ grid:
\begin{equation}
\operatorname{Grid}(X) = \frac{1}{9}\sum_{c\in\mathcal{C}} X_c.
\end{equation}
Across models, each metric is computed strictly over common valid cohorts for uncompromised fairness.

\subsection{Granularity Gains}
\label{sec:appendix_granularity}

To quantify how generative renderers benefit from structural scaffolding without confounding appearance cues, we evaluate the geometric information ladder ($P_0 \to P_1 \to P_2$) strictly under the neutral appearance condition ($A_N$).
For any metric $X \in \{\mathrm{WP}, \mathrm{RF}, \mathrm{AP}\}$, the paired marginal change between draft granularities $P_a$ and $P_b$ is defined as:
\begin{equation}
\Delta_{ab}^{X} = X_{(P_b, A_N)} - X_{(P_a, A_N)}.
\end{equation}
All comparisons enforce strictly paired world families, keeping underlying metric cohorts fixed before constructing $\mathrm{WP}$.

As summarized in \Cref{tab:main_results}, we report the net endpoint gains from abstract proxy scaffolds ($P_0$) to dense preview drafts ($P_2$) across three corresponding columns:
\begin{equation}
\begin{split}
g_s\,(\text{Structure}) = \Delta_{02}^{\mathrm{WP}}, 
\\
g_a\,(\text{Appearance}) = \Delta_{02}^{\mathrm{RF}}, 
\\
g_b\,(\text{Binding}) = \Delta_{02}^{\mathrm{AP}}.
\end{split}
\end{equation}
Differences preserve their algebraic signs: positive values signify performance gains enabled by richer structural cues, whereas negative values expose degradation or visual interference.
While \Cref{tab:main_results} focuses on endpoint dynamics $\Delta_{02}$, adjacent transitions ($\Delta_{01}$ and $\Delta_{12}$) and per-level absolute scores are documented in supplementary material to diagnose non-monotonicity and saturation behaviors.

\subsection{Conflict Robustness}
\label{sec:appendix_conflict}

To further evaluate how reliably generative renderers disentangle structural scaffolding from deceptive visual cues, we probe the model's resistance to contradictory source-target pairs. Fixing the structural scaffold at the canonical simulation operating point ($P_1$), \Cref{tab:main_results} reports four complementary metrics:
\begin{equation}
\begin{split}
\text{Adherence}\uparrow &= \mathrm{RF}_{(P_1, A_C)}, 
\\
\text{Retention}\uparrow &= \mathrm{AP}_{(P_1, A_C)}, \\
\text{Penalty}\;(\mathrm{CP})\downarrow &= \mathrm{RF}_{(P_1, A_A)} - \mathrm{RF}_{(P_1, A_C)},
\\
\text{Leakage}\;(\mathrm{SL})\downarrow &= \mathrm{SL}_{(P_1, A_C)}.
\end{split}
\end{equation}
\textit{Adherence} and \textit{Retention} measure absolute target attribute correctness and temporal binding under contradictory conditions ($A_C$).
\textit{Conflict Penalty} ($\mathrm{CP}$) quantifies the relative fidelity degradation incurred when source cues transition from aligned ($A_A$) to conflicting ($A_C$); lower values denote superior immunity to deceptive source distractors.

\vspace{-1em}
\paragraph{Source Leakage.}
$\mathrm{SL}$ explicitly tracks the erroneous persistence of conflicting source colors on target entities. For the set of verified source-color probes $\mathcal{Q}_{\mathrm{SL}}$ and their valid timestamps $\mathcal{T}_q$, it is formulated as:
\begin{equation}
\mathrm{SL} = \frac{100}{|\mathcal{Q}_{\mathrm{SL}}|} \sum_{q\in\mathcal{Q}_{\mathrm{SL}}} \frac{1}{|\mathcal{T}_q|} \sum_{t\in\mathcal{T}_q} \mathbf{1}[\hat{y}_{q,t} = \text{Yes}],
\end{equation}
where $\hat{y}_{q,t} = \text{Yes}$ indicates the active presence of the contradictory source color. Probes are constructed from audited source materials across $48$ world families with explicit color contradictions. Note that low $\mathrm{SL}$ does not trivially imply correct target realization.

\subsection{Composite Overall Metric}
\label{sec:appendix_overall}

To provide a unified ranking across diverse model paradigms, the Overall column in \Cref{tab:main_results} integrates the three diagnostic perspectives via pre-registered weights:
\begin{equation}
\mathrm{Overall} = 0.50\,R + 0.25\,G + 0.25\,C_{\mathrm{rob}},
\end{equation}
allocating $50\%$ to foundational rendering capability, $25\%$ to structural granularity responsiveness, and $25\%$ to source-target conflict disentanglement. All underlying metrics are bounded in $[0, 100]$:

\vspace{-1em}
\paragraph{Rendering Score ($R$).}
Within each condition $c \in \mathcal{C}$, we combine the macro-averaged dimensional scores using a weighted geometric mean before averaging across the $3 \times 3$ grid $R = \frac{1}{9}\sum_{c\in\mathcal{C}} S_c$:
\begin{equation}
\begin{split}
S_c = 100 \cdot (\frac{\mathrm{WP}_c}{100})^{0.40}(\frac{\mathrm{RF}_c}{100})^{0.20}(\frac{\mathrm{AP}_c}{100})^{0.30}(\frac{\mathrm{VQ}_c}{100})^{0.10}.
\end{split}
\end{equation}
The geometric mean strictly penalizes degenerate shortcuts (\eg, preserving kinematics by completely ignoring target prompts, or satisfying prompts by severely corrupting the 4D world), rewarding renderers that achieve balanced dual-correctness.

\vspace{-1em}
\paragraph{Granularity Score ($G$).}
Since individual endpoint gains $g \in \{g_s, g_a, g_b\}$ lie in $[-100, 100]$, we linearly normalize them to a standard $[0, 100]$ scale (where $50$ denotes zero net change) before computing their unweighted average:
\begin{equation}
G = \frac{1}{3}\sum_{g\in\{g_s, g_a, g_b\}} \left(50 + \frac{g}{2}\right) = 50 + \frac{g_s + g_a + g_b}{6}.
\end{equation}

\vspace{-1em}
\paragraph{Conflict Robustness Score ($C_{\mathrm{rob}}$).}
We unify the four conflict metrics into a higher-is-better index in $[0, 100]$:
\begin{equation}
\begin{split}
C_{\mathrm{rob}} = \frac{1}{4}[\mathrm{RF}_{(P_1, A_C)} + \mathrm{AP}_{(P_1, A_C)} +
\\
\big(100 - \max(\mathrm{CP}, 0)\big) + \big(100 - \mathrm{SL}_{(P_1, A_C)}\big)].
\end{split}
\end{equation}
Here, the Conflict Penalty $\mathrm{CP}$ is clipped at zero to avoid rewarding negative penalties (which arise from statistical variance rather than genuine robustness), while source leakage $\mathrm{SL}$ is inverted.

While the Overall score offers a holistic summary, the underlying gains and penalties are inherently dependent on each model's baseline operating performance. Therefore, the standalone columns in \Cref{tab:main_results} remain essential for diagnosing individual failure modes and capability trade-offs.

\subsection{Analysis of Metric-Human Alignment}

\begin{figure}[!h]
\centering
\vspace{-0.6em}
\includegraphics[width=0.54\linewidth]{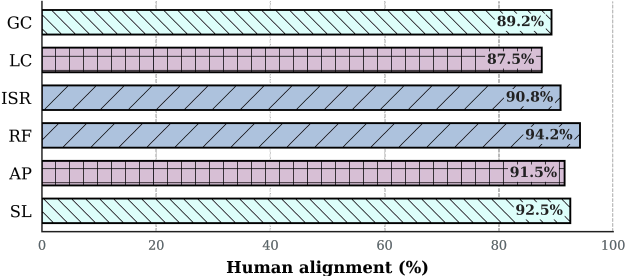}
\vspace{-0.8em}
\caption{Metric-human alignment. Agreement between automated VLM judgments and five-annotator human consensus across sampled cases, averaging $91.0\%$ across all six metrics.}
\label{fig:human_align}
\vspace{-0.6em}
\end{figure}

To validate the VLM evaluation protocol part of \ourbench, we compare algorithmic decisions against blind annotations from five independent human judges on randomly sampled test cases, using majority voting as ground truth.
As shown in \Cref{fig:human_align}, automated metrics achieve strong human agreement across all six dimensions, ranging from $87.5\%$ to $94.2\%$ (averaging $91.0\%$).
Appearance probes exhibit the highest concordance ($\mathrm{RF}: 94.2\%$, $\mathrm{SL}: 92.5\%$, $\mathrm{AP}: 91.5\%$), while physical and count metrics remain highly reliable ($\mathrm{ISR}: 90.8\%$, $\mathrm{GC}: 89.2\%$, $\mathrm{LC}: 87.5\%$).
This validates our automated benchmark suite as a faithful surrogate for human judgment.

\section{More Details of \ourmethod}

\subsection{Case Demonstration and Analysis}
\label{sec:appendix_primer_cases}

\begin{figure*}[!t]
\centering
% \vspace{-0.4em}
\includegraphics[width=0.92\linewidth]{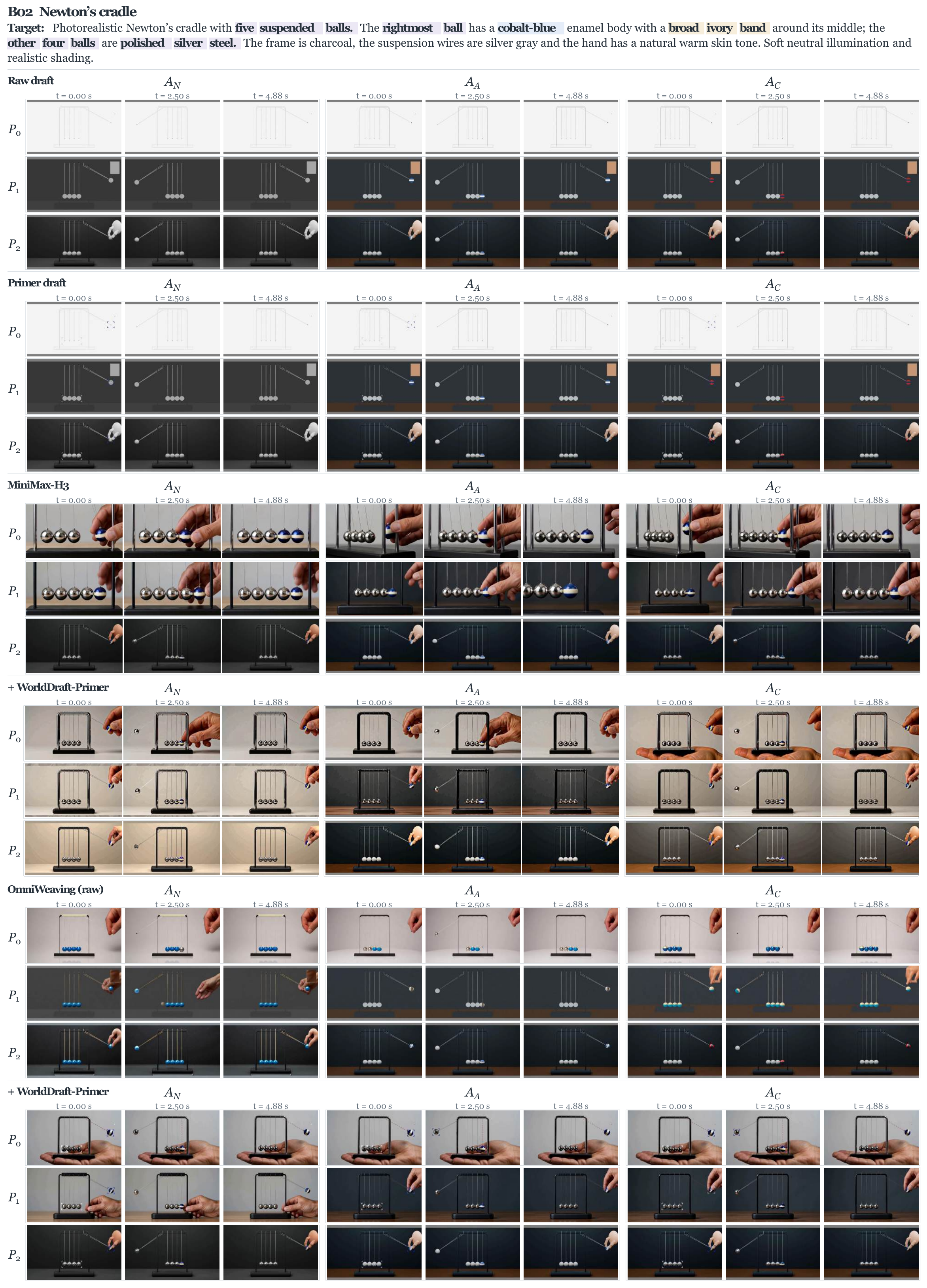}
% \vspace{-2em}
\vspace{-0.8em}
\caption{Case study on of \ourmethod.}
\label{fig:primer_1}
\vspace{-0.8em}
\end{figure*}

\begin{figure*}[!t]
\centering
% \vspace{-0.4em}
\includegraphics[width=0.92\linewidth]{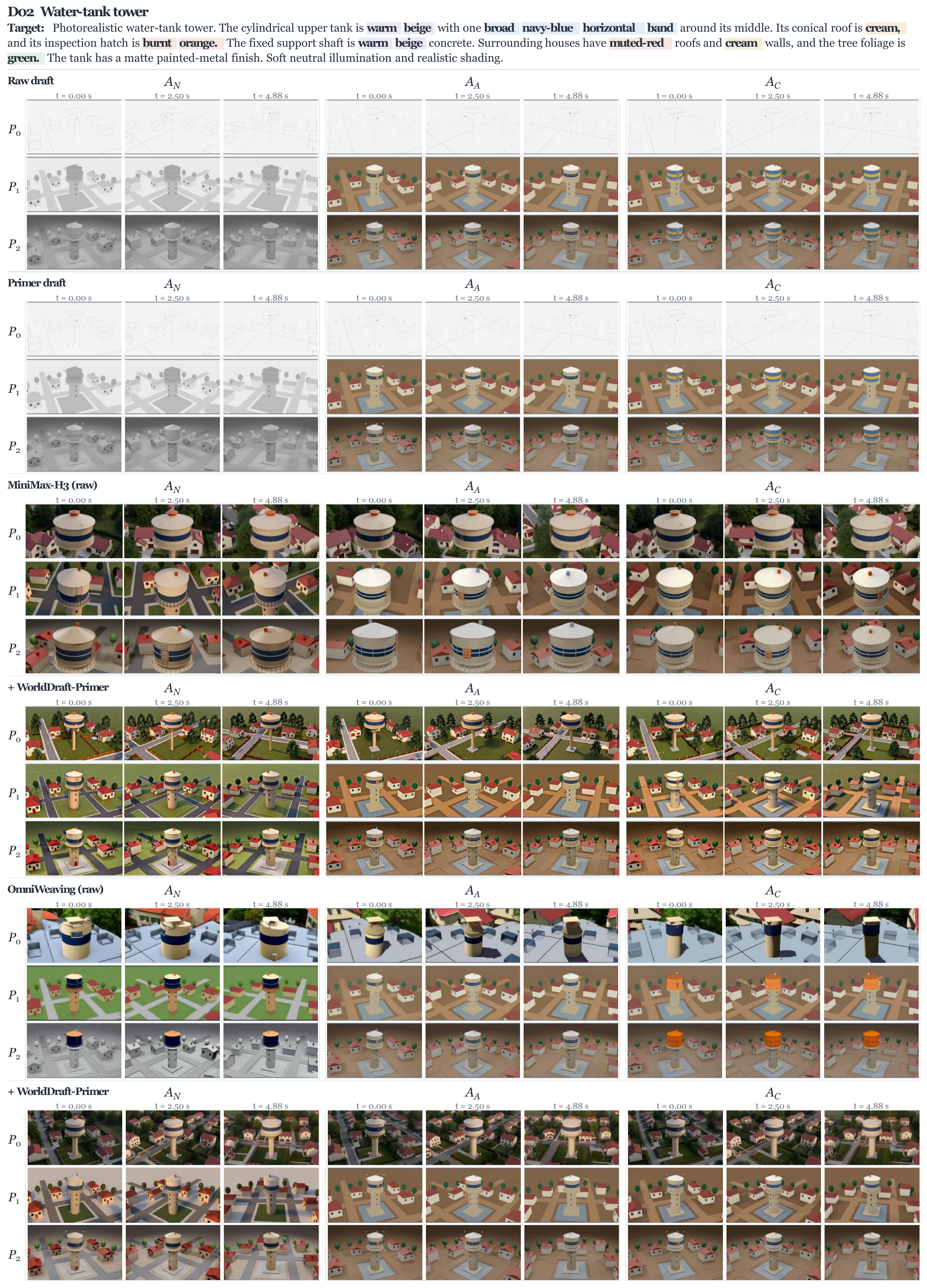}
% \vspace{-2em}
\vspace{-0.8em}
\caption{Case study on of \ourmethod.}
\label{fig:primer_2}
\vspace{-0.8em}
\end{figure*}

\Cref{fig:primer_1,fig:primer_2} visualize qualitative rollouts across the $3 \times 3$ grid for B02 (\textit{Newton's Cradle}) and D02 (\textit{Water-Tank Tower}).
Across both scenarios, \ourmethod visibly improves adherence to input drafts: it stabilizes global camera framing against erratic zoom-ins, suppresses ball dropouts in B02 while precisely isolating target colors to the rightmost sphere, and preserves surrounding town layouts in D02 while mitigating deceptive $A_C$ roof colors.

\vspace{-0.8em}
\paragraph{Capacity Dependency and Method Positioning.}
Crucially, \ourmethod operates as a lightweight visual prompting strategy whose efficacy remains inherently tethered to the underlying base model.
On weaker architectures (\eg, \omniweavinglogo~OmniWeaving), the model occasionally struggles to assimilate the injected first-frame cues into photorealistic textures, sometimes faintly echoing prompt brackets across subsequent frames.
Furthermore, under dense previs drafts ($P_2$), strong synthetic priors can still bias the renderer toward artificial, plastic-like shading.
Ultimately, \ourmethod serves as an analytical proof-of-concept demonstrating that test-time visual cues can steer frozen video models. Fully overcoming dense prior interference and enabling seamless prompt assimilation remain important challenges for future work.

\subsection{Ablation Study of \ourmethod}

To dissect the individual contributions of \textit{Object Emphasis} ($\Delta I_{\mathrm{OE}}$) and \textit{Attribute Cueing} ($\Delta I_{\mathrm{ent}} + \Delta I_{\mathrm{env}}$), we ablate each component on MiniMax-H3 across the full benchmark.
As shown in \Cref{fig:ablation}, the two mechanisms exhibit remarkable orthogonality and synergy:
\begin{figure}[!h]
\centering
\vspace{-0.6em}
\includegraphics[width=0.8\linewidth]{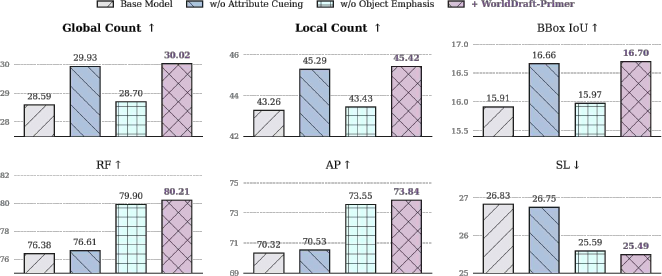}
% \vspace{-2em}
\vspace{-0.8em}
\caption{Component ablation of \ourmethod on \minimaxlogo~MiniMax-H3.}
\label{fig:ablation}
\vspace{-0.6em}
\end{figure}

Removing corner brackets (\textit{w/o Object Emphasis}) causes a sharp collapse across all spatial metrics: Global Count drops from $30.02$ to $28.70$, Local Count from $45.42$ to $43.43$, and BBox IoU from $16.70$ to $15.97$ (falling back near the base model), demonstrating that geometric brackets are indispensable for anchoring coordinates and preventing entity dropouts. Conversely, removing color and environmental hints (\textit{w/o Attribute Cueing}) cripples appearance metrics, dropping $\text{RF}$ from $80.21$ to $76.61$ and $\text{AP}$ from $73.84$ to $70.53$, while increasing Source Leakage to $26.75\%$. 
Crucially, integrating both components achieves the highest score across all six metrics simultaneously, confirming that our design decouples spatial trajectory anchoring from surface attribute binding without mutual interference.

\subsection{Text Prompt Engineering Baseline (Text PE)}
\label{sec:appendix_text_pe}

Our Text PE baseline adapts Video Prompt Optimization (VPO) \citep{cheng2025vpo} for generative rendering under four strict controls:
(\textbf{\textit{i}})~the target prompt is optimized once per world family ($p_i \to p'_i$) and shared across all nine factorial conditions to preserve paired design;
(\textbf{\textit{ii}})~the meta-prompt enriches descriptive appearance attributes while explicitly delegating kinematics, spatial layout, and camera motion to the draft;
(\textbf{\textit{iii}})~generations use identical random seeds, drafts, and model inference templates; and
(\textbf{\textit{iv}})~all evaluations strictly assess adherence to the original prompt $p_i$, preventing spurious metric shifts.

\subsection{System Prompt for Visual Prompting}
\label{sec:appendix_primer_prompt}

To automate first-frame cue generation of \ourmethod, we deploy Qwen3.5-27B \citep{yang2025qwen3} as an agentic visual grounding reasoner.
Given the target prompt and sampled draft frames, the agent is instructed to:
(\textbf{\textit{i}})~identify and ground key entity boxes in the first frame based on geometry and motion rather than conflicting source colors;
(\textbf{\textit{ii}})~extract target attributes (\eg, specified colors and patterns) from the prompt; and
(\textbf{\textit{iii}})~programmatically invoke cropping and SVG drawing tools to generate minimalist corner brackets and localized attribute cues.
The complete system prompt is detailed below:

\begin{tcolorbox}[notitle, sharp corners, breakable, colframe=BlueGreen!60, colback=gray!10, 
       boxrule=3pt, boxsep=0.5pt, enhanced, 
       shadow={3pt}{-3pt}{0pt}{opacity=1,newgray},
       title={System Prompt for \ourmethod}]\label{box:textual_loss}
       \footnotesize
       \setstretch{1}
       {\fontfamily{pcr}\selectfont
\begin{lstlisting}
"""You are a visual grounding agent for abstract draft videos. Use ONLY supplied RGB frames and original render prompt. Cuboids and rectangles can represent cars; do not require photorealistic appearance. Source colors may conflict with desired colors: identify instances by geometry, initial location and motion, not desired color. Never invent objects absent in frame zero.

The five INITIAL images come from the SAME video in chronological order: frame 0, frame 20, frame 40, frame 60, frame 80. They are successive observations, not independent scenes. Associate the same physical object across these frames using its position, shape, spatial relations and observed motion. Determine left-turning versus straight-driving roles from the observed trajectory across frames, not from a single-frame arrow, lane position or requested target color alone. Before drawing a motion-defined target, include the supporting frame indices and observed movement in its description. If the sampled frames do not establish its identity reliably, omit that target and disclose the uncertainty.

After identifying an object across time, return to ORIGINAL frame 0 to locate it. Every output box must use that object's frame-0 position; never substitute a later-frame position or add an object absent in frame 0. Subsequent tool-returned crops and drawing previews are inspection views, not additional video time steps; their order does not indicate motion. The current inspect_crop tool magnifies frame 0 only; use the five initial images for temporal evidence.

Your task is to choose a SMALL set of objects whose target color is explicitly stated in the prompt, locate them in frame 0, and place desired-color cues: outside corner markers plus sparse desired-color dashed hatching inside the object. Use tools actively: inspect_crop can magnify a region; draw_boxes renders your proposed boxes and returns both a diagnostic full-rectangle preview and the actual marker image. Inspect returned images, compare boundaries against ORIGINAL frame 0, revise if needed, then call finish. At least one draw and a subsequent visual inspection are required. Choose your own calls and corrections. Prefer tight complete object boxes, avoiding unrelated objects. Do not frame an entire machine when describing only its platform. Drawings are debugging aids, not new evidence of object identity.

All bbox coordinates are normalized [left,top,right,bottom] on the FULL ORIGINAL 832x480 image including padding, even when looking at a crop. Crops report their full-frame coordinates. Color rgb is a coarse encoding of an explicit color in the original prompt. prompt_evidence must be an exact contiguous quote binding the object to that color. Do not repaint object fills. No text/labels in the final condition. This experiment tests agentic object localization/color only; texture and environment are not verified by it.

Each draw_boxes call replaces ALL previous boxes, never appends. id is a stable short identifier; description describes the source instance and location. If a tool rejects arguments, correct them. After viewing the last actual corner result, finish with a concise explanation of what was checked and any remaining uncertainties. If you cannot localize an object reliably, omit it and disclose the omission.

HARD TOOL BUDGET: at most 8 model turns, one tool call per turn, and at most TWO successful inspect_crop calls. Each turn includes an explicit remaining-budget reminder. Prefer drawing by turn 3; spend remaining turns on useful corrections and finish. Do not keep cropping the same object. The final turn is reserved for finish when a drawing exists. You must inspect the last drawing in a subsequent turn before finish: never draw and finish in the same response. A conservative partial set with disclosed omissions is preferable to exhausting the budget without a reviewed result.

SELECTION RULES: COVER every object, component or group of objects for which the prompt states a color or a pattern and which is visible in frame 0; an omission must be justified in finish.omissions. Identify instances by geometry, initial position, source numbers and cross-frame motion, never by the desired color alone. Use one box per instance (shape=ellipse for round balls, rectangle otherwise) for up to SIXTEEN individual instances; prefer individual boxes whenever the instances are few enough (e.g. ten pendulum balls each get their own box and color). When the prompt describes MANY small similar instances (a field of dominoes, all gondolas of a Ferris wheel, rows of bottles), use ONE box with group=true around that whole group instead of skipping it; at most SIX group boxes. A group box carries the group's color in rgb; a regular arrangement such as rows alternating two colors is expressed with a texture (kind=alternating along the direction in which the rows alternate, listing the 2-4 colors in their cycle order). Patterns that live inside each small instance (e.g. a band on every domino) cannot be shown on a group box; mention them in omissions. The program draws short L-shaped corner marks just OUTSIDE each box in the desired color; a corner is dropped if it would leave the image or touch another selected box's marks, and a box needs at least two clean corners. When a box is reported as skipped, adjust it or drop it and list it in omissions. An attempted box is not an accepted annotation.

Focus the inventory on foreground instances and their visible parts. Do not use a background surface such as felt, water or asphalt as one giant object box enclosing the foreground instances; background/environment prompting is outside this run.


COLOR AND TEXTURE CUES: For each selected object the program draws (1) desired-color L-shaped corner marks outside its box and (2) sparse desired-color diagonal dashes INSIDE the object's segmented silhouette (the box is used to segment the object, so make it tight around exactly that object or component). Use separate boxes for separate components that have different colors (e.g. a platform, its base and its arms). When the prompt describes a pattern on an object, add it in `textures` instead of drawing a separate box for a thin band: kind=band for one differently colored section (e.g. "a broad ivory band around its middle" on an upright cup: axis=vertical, start=0.4, end=0.6, colors=[ivory]; a longitudinal stripe along a car: axis=short, start=0.4, end=0.6), kind=alternating for repeated stripes cycling 2-4 colors (e.g. "red-and-white stripes" on a barrier: axis=long, segments about 6-8, colors=[red, white]). axis is vertical/horizontal in the image, or long/short along the object's own silhouette. Positions are fractions of the object's extent along that axis. rgb is the object's main target color; texture colors are only the pattern colors. Every texture needs an exact prompt quote. Inspect the returned drawing: the dashes of each color must fall on the correct object and the correct part of it; revise boxes or texture parameters if not. Do not add textures that the prompt does not describe.
"""

\end{lstlisting}
}
\end{tcolorbox}

\section{Exhibition Board}
\label{sec:appendix_exhibition}

To provide an exhaustive qualitative autopsy across physical regimes while adhering to submission file size limits, complete high-resolution $3 \times 3$ grid rollouts across all six dynamic challenge categories are hosted on our interactive project page: \url{https://haroldchen19.github.io/WorldDraft-site}.
Below, we analyze representative failure cases from the top-four benchmarked models (\lynnreallogo~LynnReal-Omni, \seedancelogo~Seedance~2.5, \minimaxlogo~MiniMax-H3, and \wanlogo~Wan~3.0).
Even among these top performers, models systematically struggle to simultaneously satisfy visual plausibility, 4D structural preservation, and target attribute realization. 
These qualitative results substantiate three primary insights:

\vspace{-0.8em}
\paragraph{Appearance realism decouples from structural fidelity.}
Superficial visual quality frequently masks severe structural violations. 
In C05 (\textit{Occlusion \& Reappearance}), models synthesize photorealistic warehouse textures and forklifts, yet arbitrarily rearrange shelving aisles, alter obstacle depths, and distort occlusion orders.
Similarly, in A01 (\textit{Multi-Entity Orchestration}), \minimaxlogo~MiniMax-H3 introduces unprompted camera rotations away from the canonical top-down view, while \wanlogo~Wan~3.0 in F03 transforms dense, orderly domino arrays into sparse irregular blocks.
These cases confirm that achieving high visual quality or satisfying isolated appearance QA queries does not equate to faithful generative rendering.

\vspace{-0.8em}
\paragraph{Granularity returns are multidimensionally asymmetric.}
Increasing draft granularity does not guarantee monotonic improvements across all generative axes.
While detailed geometry ($P_2$) helps resolve local entity contours, it frequently induces composition drift, under-stylization, or degraded appearance binding.
In E03 (\textit{Multi-Stage Dynamics}), \minimaxlogo~MiniMax-H3 under $A_N$ zooms in excessively as granularity increases from $P_0$ to $P_2$, causing the umbrella to drift out of frame. 
In B05 (\textit{Multi-Stage Dynamics}), \seedancelogo~Seedance~2.5 fails to photorealistically stylize dense drafts at $P_1$ and $P_2$, retaining synthetic, toy-like block artifacts despite explicit prompts for painted steel.
Higher input granularity thus alters the model's visual attention, often trading global framing stability for local geometric detail.

\vspace{-0.8em}
\paragraph{Source leakage persists across dense scaffolds.}
Resolving contradictory appearance cues ($A_C$) requires renderers to simultaneously overwrite misleading pixels while retaining underlying kinematics.
However, models consistently treat source colors as uneditable geometry:
in C05 at $(P_1, A_C)$, \minimaxlogo~MiniMax-H3 replicates the deceptive red bodywork of the draft rather than synthesizing the target yellow;
in E03 at $(P_2, A_C)$, \seedancelogo~Seedance~2.5 faithfully tracks the complex umbrella canopy but tenaciously retains its conflicting red-and-white stripes, failing to bind target yellow and navy-blue hues.
Crucially, even when the draft provides dense, unambiguous geometric scaffolds ($P_2$), models remain vulnerable to source color leakage, underscoring the urgent need for explicit cross-modal disentanglement mechanisms.

% \begin{figure*}[!t]
% \centering
% % \vspace{-0.4em}
% \includegraphics[width=\linewidth]{figures/Figure_case_1.pdf}
% \vspace{-2em}
% % \vspace{-0.8em}
% \caption{Case study on I. Multi-Entity Orchestration of \ourbench.}
% \label{fig:case_1}
% \vspace{-0.8em}
% \end{figure*}

% \begin{figure*}[!t]
% \centering
% % \vspace{-0.4em}
% \includegraphics[width=\linewidth]{figures/Figure_case_2.pdf}
% \vspace{-2em}
% % \vspace{-0.8em}
% \caption{Case study on II. Multi-Stage Dynamics of \ourbench.}
% \label{fig:case_2}
% \vspace{-0.8em}
% \end{figure*}

% \begin{figure*}[!t]
% \centering
% % \vspace{-0.4em}
% \includegraphics[width=\linewidth]{figures/Figure_case_3.pdf}
% \vspace{-2em}
% % \vspace{-0.8em}
% \caption{Case study on III. Occlusion \& Reappearance of \ourbench.}
% \label{fig:case_3}
% \vspace{-0.8em}
% \end{figure*}

% \begin{figure*}[!t]
% \centering
% % \vspace{-0.4em}
% \includegraphics[width=\linewidth]{figures/Figure_case_4.pdf}
% \vspace{-2em}
% % \vspace{-0.8em}
% \caption{Case study on IV. Camera Motion of \ourbench.}
% \label{fig:case_4}
% \vspace{-0.8em}
% \end{figure*}

% \begin{figure*}[!t]
% \centering
% % \vspace{-0.4em}
% \includegraphics[width=\linewidth]{figures/Figure_case_5.pdf}
% \vspace{-2em}
% % \vspace{-0.8em}
% \caption{Case study on V. Non-Rigid Deformation of \ourbench.}
% \label{fig:case_5}
% \vspace{-0.8em}
% \end{figure*}

% \begin{figure*}[!t]
% \centering
% % \vspace{-0.4em}
% \includegraphics[width=\linewidth]{figures/Figure_case_6.pdf}
% \vspace{-2em}
% % \vspace{-0.8em}
% \caption{Case study on VI. Hybrid Interactions of \ourbench.}
% \label{fig:case_6}
% \vspace{-0.8em}
% \end{figure*}

\end{document}